\PassOptionsToPackage{table}{xcolor}
\documentclass[a4paper,fleqn]{cas-dc}

\usepackage[numbers,sort&compress]{natbib}

\usepackage{graphicx}
\usepackage{tikz}
\usetikzlibrary{positioning,arrows.meta}
\usepackage{multirow}
\usepackage{url}
\usepackage{algorithm}
\usepackage{algpseudocode}

\usepackage{amssymb}
\usepackage{amsmath}
\usepackage{amsthm}

\newtheorem{theorem}{Theorem}

\newtheorem{proposition}[theorem]{Proposition}

\theoremstyle{definition}

\theoremstyle{definition}

\usepackage{placeins}
\usepackage{needspace}
\usepackage{booktabs}
\usepackage{makecell}
\usepackage{tabularx}
\usepackage{float}
\usepackage{rotating}
\usepackage{pifont}
\usepackage{xcolor}
\usepackage{enumitem}
\usepackage{caption}
\usepackage{subcaption}
\usepackage{array}

\newcolumntype{C}[1]{>{\centering\arraybackslash}m{#1}}

\hypersetup{
    colorlinks=true,
    linkcolor=blue,
    citecolor=blue,
    urlcolor=blue,
    filecolor=magenta,
    bookmarks=true,
    bookmarksnumbered=true,
    bookmarksopen=true,
    pdfstartview=FitH,
    breaklinks=true,
    pdfborder={0 0 0},
pdftitle={Regularizing Surrogate-State Displacements for Multi-Step Gradient Inversion in Federated Learning},
    pdfauthor={Li Xia, Jing Yu, Zheng Liu, Sili Huang, Wei Tang, and Xuan Liu},
    pdfsubject={Federated Learning Security},
    pdfkeywords={Federated Learning, Gradient Inversion Attack, Privacy}
}

\usepackage{cleveref}
\crefname{theorem}{Theorem}{Theorems}
\crefname{lemma}{Lemma}{Lemmas}
\crefname{corollary}{Corollary}{Corollaries}
\crefname{proposition}{Proposition}{Propositions}
\crefname{definition}{Definition}{Definitions}
\crefname{equation}{Eq.}{Eqs.}
\Crefname{equation}{Equation}{Equations}
\crefname{figure}{Fig.}{Figs.}
\Crefname{figure}{Figure}{Figures}
\crefname{table}{Table}{Tables}
\Crefname{table}{Table}{Tables}
\crefname{section}{Section}{Sections}
\Crefname{section}{Section}{Sections}

\usepackage{etoolbox}
\makeatletter
\newif\ifnlsme@appendix
\pretocmd{\appendix}{\nlsme@appendixtrue}{}{}
\let\nlsme@oldseccntformat\@seccntformat
\renewcommand{\@seccntformat}[1]{%
  \ifnlsme@appendix
    \ifstrequal{#1}{section}
      {Appendix~\csname the#1\endcsname.\quad}
      {\nlsme@oldseccntformat{#1}}%
  \else
    \nlsme@oldseccntformat{#1}%
  \fi
}
\makeatother

\ExplSyntaxOn
\RenewDocumentCommand \printorcid { } { }
\RenewDocumentCommand \printemails { }
{
  \group_begin:
  \int_compare:nNnTF { \int_use:N \g_ead_int } > { 0 }
  {
    \tex_let:D \thefootnote \relax
    \footnotetext
    {
      \raggedright
      \int_compare:nTF { \g_ead_int = 1 }
      { \textit{E-mail~address:\c_space_token} }
      { \textit{E-mail~addresses:\c_space_token} }
      \seq_use:Nn \g_stm_ead_seq { ;~ }
    }
  }
  { }
  \group_end:
}
\ExplSyntaxOff

\begin{document}
\let\WriteBookmarks\relax
\def\floatpagepagefraction{1}
\def\textpagefraction{.001}

\shorttitle{}
\shortauthors{Xia et~al.}

\title [mode = title]{NL-SME: Enhancing Multi-Step Gradient Inversion through Adaptive Two-State Matching}

\author[1]{Li Xia}
\ead{xiali@muc.edu.cn}

\author[1]{Jing Yu}
\cormark[1]
\ead{jing.yu@muc.edu.cn}

\author[1]{Zheng Liu}
\ead{liuzheng@muc.edu.cn}

\author[1]{Sili Huang}
\ead{huangsili@muc.edu.cn}

\author[1]{Wei Tang}
\ead{tangocean@bupt.cn}

\author[1]{Xuan Liu}
\cormark[1]
\ead{liuxuan@muc.edu.cn}

\affiliation[1]{
organization={Key Laboratory of Ethnic Language Intelligent Analysis and Security Governance of MOE, Minzu University of China},
city={Beijing},
postcode={100081},
country={China}
}

\cortext[1]{Corresponding authors: Jing Yu and Xuan Liu.}

\begin{abstract}
Federated learning (FL) enables collaborative model training among multiple clients without directly sharing their raw data.
However, gradient inversion attacks can exploit the model updates exchanged during training to reconstruct clients' private data.
With multiple local training steps, an observed update combines gradients evaluated at changing model states and potentially different mini-batches, while intermediate states remain hidden.
Restricting gradient evaluation to a single state on the segment between the initial and final models can exclude states that better fit the accumulated update.
We consider reconstruction from an individual client's endpoint models, assuming that the training labels are known.
We propose NL-SME, a regularized two-state gradient surrogate for matching multi-step client updates without replaying local training.
The surrogate adapts its evaluation center, regularizes its displacement, and averages raw parameter gradients at two shared symmetric states.
We evaluate NL-SME against existing gradient inversion methods on FEMNIST and CIFAR-100 with varying numbers of local training steps.
The results show improved reconstruction quality, highlighting the privacy risks of sharing individual client updates after multiple local steps.
\end{abstract}


\maketitle

\section{Introduction}

Federated learning (FL) enables multiple clients to collaboratively train a shared model through the exchange of locally computed updates~\cite{kairouz2021advances,li2020federated,mothukuri2021survey}.
By combining information from distributed data sources, this framework supports model training without collecting clients' raw data at a central server.
In Federated Averaging (FedAvg), clients perform multiple local stochastic gradient descent (SGD) steps before uploading an update, reducing communication frequency and keeping intermediate computations local~\cite{mcmahan2017communication,li2020federated}.

However, keeping data local does not prevent shared gradients or model updates from revealing private information~\cite{gosselin2022privacy,carletti2025sok}.
Gradient inversion attacks exploit this leakage by reconstructing training samples from the observed optimization signals~\cite{geiping2020inverting,zhu2026cracking,guo2026exploring}.
These risks have motivated defenses that modify local training or protect individual gradients during aggregation~\cite{gao2024defending,dong2025mtsec,zhang2026secure}.
For multi-step updates, these signals combine gradients evaluated at changing model states and potentially different mini-batches~\cite{dimitrov2022data,zhu2023surrogate}.
We study this reconstruction problem for an honest-but-curious server that observes an individual client's initial and final models within one communication round, with the training labels assumed known.

Early gradient inversion methods formulate reconstruction as an optimization problem, treating candidate images as variables and minimizing the discrepancy between candidate and observed gradients~\cite{zhu2019deep,zhao2020idlg,geiping2020inverting}.
Subsequent studies improve reconstruction through label inference, alternative matching objectives, partial gradient matching, and image or gradient regularization~\cite{zhao2020idlg,geiping2020inverting,fan2025boosting,wan2026beyondsgd}.
Generative and architectural priors offer another way to constrain the reconstruction space~\cite{wu2025dggi,yu2025ginas,pu2025cggl}.
For multi-step updates, however, reconstruction must also account for gradients accumulated during local training.
Simulation-based approaches address this by reproducing local optimization on candidate data and matching the resulting model update~\cite{dimitrov2022data,geiping2020inverting}.
The computational cost grows with the number of simulated steps.
A surrogate approach instead evaluates candidate-data gradients at a learned state on the segment connecting the observed endpoint models, avoiding explicit simulation of the full sequence~\cite{zhu2023surrogate}.
This raises a question: how can a compact gradient surrogate account for variation across model states when reconstructing private images from multi-step updates?

The observed endpoints determine the net parameter change but do not directly identify a representative state for evaluating candidate-data gradients.
For fixed candidate data, restricting the evaluation location to a prescribed region can exclude states that provide a better fit to the update.
Allowing the evaluation state to leave this region introduces additional parameters, motivating regularization of its displacement.
Moreover, when the gradient map is nonlinear, its value at a representative state need not equal the average of gradients across states.
These considerations motivate regularized state adaptation and gradient averaging over a small set of evaluation states.

We propose NL-SME, a regularized two-state gradient surrogate for reconstructing private images from multi-step model updates.
Specifically, we allow the evaluation center to depart from the endpoint segment while penalizing the actual state displacement.
To account for gradient variation across states, we use a symmetric pair around this center, shared by all candidate images, and average the two raw parameter gradients before matching the observed update.
The resulting formulation jointly optimizes candidate images, the evaluation center, and the shared state separation without replaying the client's local training.
Component ablations show that regularized center displacement and shared two-state gradient averaging both improve reconstruction quality.
We also consider different local-training lengths and evaluate reconstruction on FEMNIST and CIFAR-100.

Our main contributions are as follows:
\begin{itemize}
    \item We propose NL-SME, which adapts the evaluation center, regularizes its displacement, and averages raw parameter gradients at two shared symmetric states.
    It requires two matching parameter-gradient evaluations per iteration without replaying local training.

    \item We establish sufficient conditions for center displacement to improve the regularized fitting objective with fixed candidate images and evaluation width, and characterize two-state averaging through a local expansion.

    \item Experiments on FEMNIST and CIFAR-100 demonstrate improved reconstruction from multi-step updates, with ablations supporting the two components.
    Under the default CIFAR-100 setting, NL-SME improves mean PSNR over SME by $4.65$ dB with $2000$ matching parameter-gradient evaluations per method.
\end{itemize}

\section{Background and Related Work}
\label{sec:background}

\subsection{Federated Learning}
\label{subsec:federated_learning}

The server broadcasts the current model parameters to a subset of clients for each round of FedAvg.
Each client subsequently samples mini-batches from its local dataset to compute local gradients and update its model.
To alleviate the communication burden, selected clients update their local models multiple times using SGD, followed by sending the resulting model parameters to the server.
The server then averages these uploaded parameters in proportion to the clients' local sample counts to form a new global model~\cite{mcmahan2017communication,li2020federated}.

Notably, in this strategy, the server can deduce a client's accumulated update by comparing the parameters of the old global model and the model uploaded by that client.
However, this strategy combines gradients computed at successive model states on potentially different mini-batches.
The computed gradients and intermediate model states are not sent back to the server.
Such accumulation can complicate data reconstruction because the update direction need not match that of a gradient computed at the old global model~\cite{dimitrov2022data,zhu2023surrogate}.

\subsection{Gradient Leakage Attack}
\label{subsec:gradient_leakage_attack}

DLG iteratively adjusts dummy inputs and labels to bring their generated gradients closer to the observed gradients~\cite{zhu2019deep}.
Zhao et al.~\cite{zhao2020idlg} discovered that labels can be inferred from individual-sample gradients under cross-entropy training with one-hot targets, enabling reconstruction with fixed labels.
Geiping et al.~\cite{geiping2020inverting} adopted cosine similarity for gradient matching and introduced total variation to regularize the reconstructed images.
Fan et al.~\cite{fan2025boosting} introduced partial gradient matching and gradient regularization in FEDLEAK to improve reconstruction.

Other studies have improved reconstruction by incorporating additional information and prior knowledge.
Yin et al.~\cite{yin2021see} assumed that batch-normalization statistics are accessible, which provide additional cues to regularize the reconstructed images.
Wu et al.~\cite{wu2025dggi} proposed DGGI to enhance reconstruction by incorporating a pretrained diffusion model as an image prior.
GI-NAS uses the initial gradient-matching loss to select a reconstruction network, allowing the architectural prior to vary across target batches~\cite{yu2025ginas}.
Pu et al.~\cite{pu2025cggl} proposed CGGL to reconstruct private images using diffusion priors, assuming that malicious clients can poison uploaded updates.
Nonetheless, some studies have criticized assumptions of access to private batch statistics in practical FL settings~\cite{huang2021evaluating,du2025sok}.

Simulation-based methods optimize candidate data through local-training simulations to produce updates that match the observed multi-step update~\cite{dimitrov2022data}.
Zhu et al.~\cite{zhu2023surrogate} proposed SME to match gradients at a learned state interpolated between the endpoint models, allowing data reconstruction without simulating local training.

\subsection{Gradient Leakage Defense}
\label{subsec:gradient_leakage_defense}

Defenses against gradient leakage can be broadly classified into cryptography-based and perturbation-based methods.
Cryptography-based methods conceal individual client updates through secure aggregation or encrypted computation~\cite{MPC2,HE2}.
This prevents direct application of attacks that require an individual client's update when only an aggregate is observable and individual contributions cannot be isolated.
However, these methods incur additional computational and communication costs.
Recent schemes seek to reduce these costs through lightweight homomorphic encryption, functional encryption, and decentralized functional encryption~\cite{zhang2025parfs,yu2025lightweight,guan2025pefed}.
Verifiable FL schemes combine protection of client updates with verification of aggregation results~\cite{niu2025verifiable,ma2025pevfl}.

Perturbation-based methods seek to limit the information available for reconstruction by modifying shared gradients or model updates.
This strategy, however, involves a trade-off between utility and privacy that depends on the perturbation strength and training protocol~\cite{carletti2025sok}.
Client-level differentially private FL can combine update clipping, calibrated noise, and privacy accounting across training rounds~\cite{geyer2017differentially}.
Gradient sparsification retains large-magnitude elements to reduce communication, and gradient quantization represents gradient values with fewer bits~\cite{aji2017sparse,du2025sok}.
Recent schemes combine gradient compression with adaptive noise injection or shuffling to balance privacy, utility, and communication cost~\cite{shen2026flapgc,xiao2025communication}.
Dedicated defenses include Soteria, which perturbs data representations, and Guardian, which optimizes transformed gradients to balance privacy protection and model performance~\cite{soteria,guardian}.
Recent studies also address reconstruction risks in federated fuzzy models and protect user representations in federated cross-domain recommendation~\cite{yin2026trilevel,do2025federated}.

\section{Preliminaries}
\label{sec:revisit}

\subsection{Threat Model}
\label{subsec:threat_model}
\label{par:attack_setting}

The threat model concerns privacy leakage from an individual client's multi-step update in horizontal FL. It follows the honest-but-curious setting examined in recent studies of gradient inversion~\cite{du2025sok}.

\textbf{Envisioned attacker(s).}
The adversary is an honest-but-curious server that targets one client while following the prescribed FL protocol. The server observes the client's uploaded model without modifying the architecture, loss, or local training procedure.

Firstly, Goal and Methodology.
The attacker's goal is to reconstruct the private images $X$ used for local training. Its available signal is the endpoint difference $u=w_0-w_T$, and the reconstruction problem is conditioned on the target labels $y$.

Secondly, Timing.
The attack uses a single observed communication round. Starting from the received model $w_0$, the client performs $T$ local SGD steps and returns $w_T$. The server knows both endpoints, but the intermediate states $\{w_\tau\}_{\tau=1}^{T-1}$ and per-step gradients $\{g_\tau\}_{\tau=0}^{T-1}$ remain hidden. No observations from other rounds are used.

Thirdly, Abilities and Knowledge.
The attacker knows the architecture, loss, local dataset size $N$, image shape, normalization, and local training configuration. The latter includes the learning rate $\eta_{\mathrm{tr}}$, batch size $B$, and number of local steps $T$. The sample count is available for aggregation in sample-size-weighted FedAvg~\cite{mcmahan2017communication}. The mini-batch order remains unknown.

The target labels $y$ are assumed known and remain fixed during image reconstruction, as in a recent evaluation of gradient inversion~\cite{guo2026exploring}. Label recovery from multi-step local updates has been studied under additional assumptions, including access to auxiliary data~\cite{chen2024recovering}. This does not imply that $y$ can always be recovered from $u$. The reconstruction problem is therefore conditional on label availability.

\textbf{Threat surface.}
The leakage channel is the individual update $u$, which accumulates information from gradients computed on private data. The attack requires this update to be visible before aggregation. Settings in which secure aggregation prevents access to individual updates fall outside this threat model.

\textbf{Scope.}
The setting covers image-classification tasks trained by local SGD with a constant learning rate, without momentum or weight decay. The forward computation uses evaluation mode during both local optimization and reconstruction, keeping any normalization buffers at their known pre-round values. No private batch statistics are supplied. The analysis below concerns unperturbed updates and does not assume access to the hidden training trajectory.

\subsection{Problem Formulation}
\label{subsec:problem_formulation}

For the sake of discussion, let us consider a model $f_w$ at a known state $w$ and a private dataset $\mathcal D=\{(x_i,y_i)\}_{i=1}^{N}$, consisting of images $X$ and labels $y$.
We assume an exact-gradient reference setting in which the observed gradient $\hat{g}$ is identical to $g_X(w)=\nabla_w\ell(w;X,y)$, where $\ell$ denotes the mean cross-entropy loss over $\mathcal D$.
We also assume that the labels $y$ are known and remain fixed during reconstruction.
Under these assumptions, the gradient matching problem is formulated as minimizing $\operatorname{dist}(g_{\widehat{X}}(w),\hat{g})+\lambda_{\mathrm{img}}\mathcal R_{\mathrm{img}}(\widehat{X})$ with respect to $\widehat{X}$~\cite{zhu2019deep,geiping2020inverting}.
Here, $g_{\widehat{X}}(w)=\nabla_w\ell(w;\widehat{X},y)$ is the gradient generated by candidate images $\widehat{X}$, and $\operatorname{dist}(\cdot,\cdot)$ denotes a gradient distance.
The image prior $\mathcal R_{\mathrm{img}}$ is weighted by $\lambda_{\mathrm{img}}\ge0$.
In the multi-step setting, the server instead observes an accumulated update, whose relation to single-state gradients is examined below.

\textbf{The observed multi-step update.}
Under the threat model in Section~\ref{subsec:threat_model}, the server observes $u=w_0-w_T$ after $T$ local SGD steps.
We use $F_{\mathcal A}$ to denote the mean cross-entropy loss on sample set $\mathcal A$, with $\nabla F_{\mathcal D}=g_X$.
The update rule at local step $\tau$ is given by $w_{\tau+1}=w_\tau-\eta_{\mathrm{tr}}g_\tau$, where $g_\tau=\nabla F_{\mathcal B_\tau}(w_\tau)$ is the gradient generated by mini-batch $\mathcal B_\tau$ at state $w_\tau$.
Let $\xi_\tau=g_\tau-g_X(w_\tau)$ denote the difference between the mini-batch and full-data gradients at $w_\tau$, and let $\bar\xi=T^{-1}\sum_{\tau=0}^{T-1}\xi_\tau$ denote its average over the local steps.
With a constant learning rate $\eta_{\mathrm{tr}}>0$, substituting these definitions into the sum of the local updates yields
\begin{equation}
\begin{aligned}
\bar u=\frac{w_0-w_T}{\eta_{\mathrm{tr}}T}
&=\frac1T\sum_{\tau=0}^{T-1}g_\tau\\
&=\frac1T\sum_{\tau=0}^{T-1}g_X(w_\tau)+\bar\xi.
\end{aligned}
\label{eq:full_gradient_average}
\end{equation}
Equation~\eqref{eq:full_gradient_average} expresses the normalized update as an average of full-data gradients across local model states, together with a realized mini-batch residual.
Full-batch training gives $\bar\xi=0$, but a complete mini-batch pass need not make this residual vanish because the gradients are evaluated at changing model states.
Thus, observing $u$ does not in general provide an exact gradient at one known model state for the matching problem above.

\textbf{Examining a single-state approximation.}
Let us examine the matching problem by approximating $\bar u$ with a gradient at a single state $\widetilde w$.
We use $\delta_{\mathrm{img}}(\widetilde w)=g_{\widehat{X}}(\widetilde w)-g_X(\widetilde w)$ and $\delta_{\mathrm{state}}(\widetilde w)=g_X(\widetilde w)-\bar u$ to denote the image and state components of the matching residual.
Adding and subtracting $g_X(\widetilde w)$ yields $g_{\widehat{X}}(\widetilde w)-\bar u=\delta_{\mathrm{img}}(\widetilde w)+\delta_{\mathrm{state}}(\widetilde w)$.
Here, $\delta_{\mathrm{img}}$ is the parameter-gradient difference caused by replacing the private images with candidate images at the same state.
The term $\delta_{\mathrm{state}}$ measures the discrepancy between $g_X(\widetilde w)$ and $\bar u$, including the effect of $\bar\xi$.
If $\widehat{X}=X$, $\delta_{\mathrm{img}}$ vanishes, but $\delta_{\mathrm{state}}$ need not vanish.
The following theorem gives an upper bound on $\|\delta_{\mathrm{state}}(\widetilde w)\|_2$ under a Lipschitz condition on $g_X$.

\begin{theorem}[Single-state approximation]
\label{thm:representative_state_mismatch}
\upshape
Consider the state residual $\delta_{\mathrm{state}}(\widetilde w)$ at a representative state $\widetilde w$ for the $T$-step local-SGD process above.
If $g_X=\nabla F_{\mathcal D}$ is $L$-Lipschitz on a region containing $\widetilde w$ and the local iterates $\{w_\tau\}_{\tau=0}^{T-1}$, then it holds that $\|\delta_{\mathrm{state}}(\widetilde w)\|_2\le L\varepsilon_{\mathrm{state}}(\widetilde w)+\|\bar\xi\|_2$, where $\varepsilon_{\mathrm{state}}(\widetilde w)=T^{-1}\sum_{\tau=0}^{T-1}\|w_\tau-\widetilde w\|_2$ is the mean distance from $\widetilde w$ to these iterates.
\end{theorem}

Substituting Eq.~\eqref{eq:full_gradient_average} into the definition of $\delta_{\mathrm{state}}(\widetilde w)$ yields
\begin{equation}
\begin{aligned}
&\|\delta_{\mathrm{state}}(\widetilde w)\|_2\\
&\quad=\left\|\frac1T\sum_{\tau=0}^{T-1}
 [g_X(\widetilde w)-g_X(w_\tau)]-\bar\xi\right\|_2\\
&\quad\le\frac1T\sum_{\tau=0}^{T-1}
 \|g_X(\widetilde w)-g_X(w_\tau)\|_2+\|\bar\xi\|_2\\
&\quad\le L\varepsilon_{\mathrm{state}}(\widetilde w)+\|\bar\xi\|_2.
\end{aligned}
\label{eq:single_state_mismatch}
\end{equation}
Here, the two inequalities follow from the triangle inequality and the Lipschitz condition, respectively.
For fixed $L$ and $\bar\xi$, a smaller mean distance can tighten this upper bound.
The evaluation location therefore enters the bound even when $\widehat{X}=X$, although the hidden distances in $\varepsilon_{\mathrm{state}}(\widetilde w)$ are unavailable to the attacker.
The complete proof is given in Appendix~\ref{appendix:supporting_theory}.

To relate this bound to cosine matching, let $a=g_X(\widetilde w)$ and $b=\bar u$, and assume both are nonzero.
Since $u=\eta_{\mathrm{tr}}Tb$ is a positive multiple of $b$, $u$ and $b$ have the same direction.
Then, expanding $\|a-b\|_2^2$ yields
\begin{equation}
\begin{aligned}
&1-\frac{\langle a,u\rangle}{\|a\|_2\|u\|_2}\\
&\quad=\frac{\|a-b\|_2^2-(\|a\|_2-\|b\|_2)^2}
 {2\|a\|_2\|b\|_2}\\
&\quad\le\frac{[L\varepsilon_{\mathrm{state}}(\widetilde w)+\|\bar\xi\|_2]^2}
 {2\|a\|_2\|b\|_2}.
\end{aligned}
\label{eq:single_state_direction}
\end{equation}
Here, the inequality follows by dropping the nonnegative squared difference between the norms and applying Eq.~\eqref{eq:single_state_mismatch}.
This indicates that the cosine discrepancy is small when the square of the state-residual bound is small relative to the product of the gradient norms.
However, this bound concerns the gradient $g_X(\widetilde w)$ evaluated on the true images at a single state and does not guarantee recovery of $X$ from a small cosine loss.

\textbf{From evaluation location to gradient averaging.}
To further examine the single-state approximation, let $\mu=T^{-1}\sum_{\tau=0}^{T-1}w_\tau$ be the mean of the states before the local updates.
For a nonlinear $g_X$, the identity $T^{-1}\sum_{\tau=0}^{T-1}g_X(w_\tau)=g_X(\mu)$ need not hold.
Here, the left-hand side averages gradients evaluated at the individual states, while the right-hand side evaluates the gradient at their mean.
The mean state alone does not specify how the local states are spread around it, which can affect their average gradient.
For an affine gradient map, the identity holds, although the normalized observation in Eq.~\eqref{eq:full_gradient_average} still includes $\bar\xi$.
Thus, choosing an evaluation location and accounting for gradient variation across states are distinct issues in approximating a multi-step update.

To define an observable reference for the evaluation location, consider the endpoint segment $w_{\mathrm{lin}}(t)=(1-t)w_0+t w_T$, with $t\in[0,1]$~\cite{zhu2023surrogate}.
Since the endpoints $w_0,w_T$ are observed, the segment can be constructed without access to the intermediate local states.
However, the gradients driving local updates can change direction between steps, so the segment need not contain the hidden iterates $w_\tau$.

\Needspace{6\baselineskip}
\textbf{Remark.}
The upper bounds in this section rely on the Lipschitz condition on the stated region, which may not hold across ReLU activation boundaries.
The quantities $\mu$, $\varepsilon_{\mathrm{state}}$, and $\bar\xi$ are used only for analysis and are not observed by the attacker, as they depend on hidden local states and gradients.
For the unperturbed update, these bounds concern gradient discrepancies and neither identify a true intermediate state nor establish a bound on the image error $\|\widehat X-X\|_2$.
As upper bounds, they do not rule out $\delta_{\mathrm{state}}(\widetilde w)=0$ at some state.
Additionally, an additive perturbation of the uploaded update, $u_{\mathrm{obs}}=u+\zeta$, changes the normalized observation by $\zeta/(\eta_{\mathrm{tr}}T)$.

\Needspace{24\baselineskip}

\section{Method}
\label{sec:method}

\subsection{Overview}
\label{subsec:overview}

\begin{figure*}[t] \centering
\includegraphics[width=\textwidth]{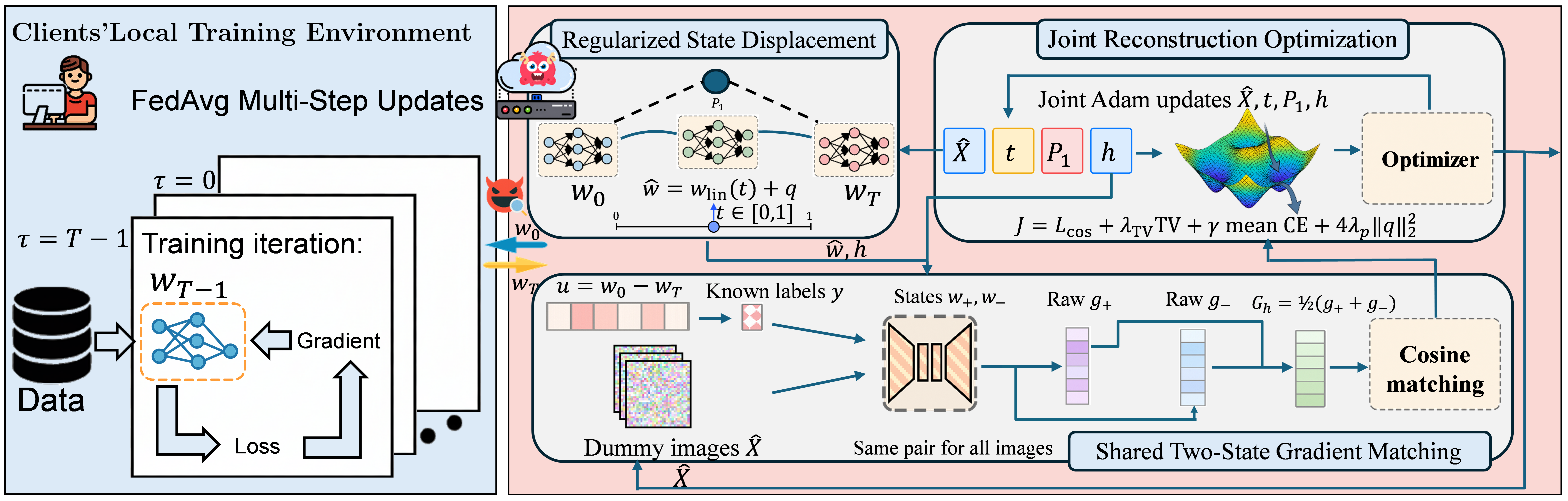}
\caption{Overview of our proposed NL-SME attack. With known labels, NL-SME reconstructs client data by matching the observed multi-step update to a regularized two-state gradient surrogate. We adapt the gradient evaluation center through regularized displacement from the endpoint segment. We average the raw parameter gradients at two symmetric states shared by all candidate images before cosine matching with the observed update.} \label{fig:illustration}
\end{figure*}

NL-SME matches the observed multi-step update with a regularized two-state gradient surrogate, jointly optimizing its evaluation center and state separation with the candidate images.
The evaluation center $\hat{w}$ can move beyond the endpoint segment through a learnable displacement $q$.
The penalty acts on the actual state displacement, controlling departures from this reference.
Moreover, to account for gradient variation across states, we evaluate the candidate images $\widehat{X}$ at a symmetric pair of model states around $\hat{w}$, with the same pair shared by all images.
The two raw parameter gradients are averaged before a single cosine comparison with the observed update $u$.
Forming this surrogate requires two parameter-gradient evaluations per iteration without replaying the client's local training sequence.
We differentiate the reconstruction objective, including image and state regularization, to jointly update $\widehat{X}$ and the state parameters with Adam.
Algorithm~\ref{alg:nlsme} summarizes the reconstruction procedure, and Figure~\ref{fig:illustration} shows how the two components enter the same objective.

\begin{algorithm}[!t]
\caption{Our attack: NL-SME}
\label{alg:nlsme}
\begin{algorithmic}[1]
\raggedright
\algrenewcommand{\algorithmiccomment}[1]{\quad // #1}
\Require $w_0,w_T$: endpoint models ($w_0\ne w_T$); $y$: known labels; $K$: maximum attack iterations.
\Ensure $\widehat{X}^*$: the reconstructed images.
\State $u\gets w_0-w_T$, $v\gets w_T-w_0$
\State $\widehat{X}\sim\mathcal{N}(0,I)$, $\widehat{X}\gets\Pi_{\mathcal{X}}(\widehat{X})$
\State $t\gets1/2$, $P_1\gets(w_0+w_T)/2$, $h\gets h_0$
\For{$k\gets0$ to $K-1$}
  \State Form $\hat{w}$, then $w_+\gets\hat{w}+hv$, $w_-\gets\hat{w}-hv$
  \State $G_h\gets\tfrac12[g_{\widehat{X}}(w_+)+g_{\widehat{X}}(w_-)]$ \Comment{raw gradients}
  \State Compute $\mathcal{J}$ by Eq.~\eqref{eq:method-objective}
  \State Update $\widehat{X},t,P_1,h$ with Adam \Comment{separate optimizers, all gradients computed before updates}
  \State Project $\widehat{X},t,h$ onto feasible domains
\EndFor
\State \Return $\widehat{X}^*=\operatorname{Denormalize}(\widehat{X})$
\end{algorithmic}
\end{algorithm}

\subsection{Regularized Surrogate-State Displacement}
\label{subsec:state_displacement}

The core idea is to allow the gradient evaluation center to move beyond the endpoint segment while penalizing its actual displacement.
Let $\hat{w}=w_{\mathrm{lin}}(t)+q$, where $w_{\mathrm{lin}}(t)$ is the endpoint reference~\cite{zhu2023surrogate} and $q$ is a learnable displacement.
To isolate center adaptation in the reconstruction objective, fix $\widehat{X},y$ and a width $h\in[0,1/2]$.
For a center $w$, consider the shared evaluation states $w+hv$ and $w-hv$, where $v=w_T-w_0$, as formalized in Section~\ref{subsec:shared_state_matching}.
Let us define $\Phi_h(w)$ as the cosine distance between their averaged raw parameter gradient and $u$, plus $\gamma$ times the mean cross-entropy at the same states.
The regularized center objective can then be written as $H_{t,h}(q)=\Phi_h(w_{\mathrm{lin}}(t)+q)+R_{\mathrm{disp}}(q)$, where $R_{\mathrm{disp}}(q)=4\lambda_p\|q\|_2^2$ and $\lambda_p>0$.
When $q=0$, the problem reduces to optimizing $t$ along the endpoint segment at the same fixed width.
The following theorem identifies conditions under which a displaced center achieves a lower regularized objective value than that attained by the best center on this segment.

\begin{theorem}[Benefit of regularized displacement]
\label{thm:regularized_center_adaptation}
\upshape
Consider the regularized center objective for fixed $\widehat{X},y,h$, $\lambda_p>0$, and $v\ne0$, with the same images, labels, and width used in both compared surrogates.
Suppose $\Phi_h$ is defined on the endpoint segment and attains its minimum over this segment at $w_{\mathrm{seg}}=w_{\mathrm{lin}}(t_{\mathrm{seg}})$, with $t_{\mathrm{seg}}\in(0,1)$.
Assume that $\Phi_h$ has an $L_\Phi$-Lipschitz gradient on a convex neighborhood of $w_{\mathrm{seg}}$.
If $\nabla\Phi_h(w_{\mathrm{seg}})\ne0$, then there exists a displacement $q$ such that the center $w_{\mathrm{seg}}+q$ lies outside the endpoint line and satisfies $H_{t_{\mathrm{seg}},h}(q)<\min_{t\in[0,1]}H_{t,h}(0)$.
\end{theorem}

We demonstrate the improvement in Theorem~\ref{thm:regularized_center_adaptation} by constructing a displacement after optimizing $t$ along the endpoint segment.
Let $d_\Phi=\nabla_w\Phi_h(w_{\mathrm{seg}})$ be the nonzero center gradient at the interior optimum.
Since the derivative along the segment vanishes at this point, it follows that $\langle d_\Phi,v\rangle=0$.
By setting $q_\alpha=-\alpha d_\Phi$ with $0<\alpha\le(L_\Phi+8\lambda_p)^{-1}$ small enough that $w_{\mathrm{seg}}+q_\alpha$ remains in the stated convex neighborhood, we obtain a nonzero displacement orthogonal to the endpoint line.
Applying the smoothness bound for $\Phi_h$ and adding the displacement penalty yields
\begin{equation}
\begin{aligned}
&H_{t_{\mathrm{seg}},h}(q_\alpha)-H_{t_{\mathrm{seg}},h}(0)\\
&\quad\le\langle d_\Phi,q_\alpha\rangle
  +\frac{L_\Phi}{2}\|q_\alpha\|_2^2+4\lambda_p\|q_\alpha\|_2^2\\
&\quad=-\alpha\|d_\Phi\|_2^2
  +\frac{\alpha^2}{2}(L_\Phi+8\lambda_p)\|d_\Phi\|_2^2.
\end{aligned}
\label{eq:method-center-decrease}
\end{equation}
Using $H_{t_{\mathrm{seg}},h}(0)=\min_{t\in[0,1]}H_{t,h}(0)$, we rearrange Eq.~\eqref{eq:method-center-decrease} and apply the same step-size bound to obtain
\begin{equation}
\begin{aligned}
&\min_{t\in[0,1]}H_{t,h}(0)-H_{t_{\mathrm{seg}},h}(q_\alpha)\\
&\quad\ge\alpha\left[1-\frac{\alpha}{2}(L_\Phi+8\lambda_p)\right]\|d_\Phi\|_2^2\\
&\quad\ge\frac{\alpha}{2}\|d_\Phi\|_2^2>0.
\end{aligned}
\label{eq:method-center-gap}
\end{equation}
Here, the step-size bound makes the factor in brackets at least one half.
Since the step size is positive and $d_\Phi\ne0$, the gap in Eq.~\eqref{eq:method-center-gap} is strictly positive.
Thus, even with the displacement penalty, the constructed center attains a lower objective value than the best reference on the endpoint segment, as stated in Theorem~\ref{thm:regularized_center_adaptation}.
The full proof is given in Appendix~\ref{appendix:state_displacement}.

The analysis above relies on keeping $\widehat{X},y,h$ fixed in both surrogates to isolate the effect of center adaptation.
When $h>0$, the reference whose center is restricted to the endpoint segment uses the same two-state averaging scheme.
Under the stated conditions, the theorem establishes the existence of a displacement that lowers the regularized fitting objective.
In practice, Adam jointly updates the candidate images, center parameters, and shared width while keeping the labels fixed, so the constructed $q_\alpha$ does not prescribe these updates.

To parameterize the center, we introduce a learnable control point $P_1$ and define its offset from the endpoint midpoint as $r=P_1-(w_0+w_T)/2$.
With $q=2t(1-t)r$, $t$ selects the reference position and scales the displacement induced by $P_1$.
For any fixed $t\in(0,1)$, every $q$ can be realized by a corresponding $P_1$, including the displacement constructed in Theorem~\ref{thm:regularized_center_adaptation}.
When $t=0$ or $t=1$, $q=0$ and the center $\hat{w}$ coincides with the corresponding endpoint.
The penalty acts on $\|q\|_2^2$, the squared displacement of the center from its selected reference on the endpoint segment.
A penalty $\lambda_p\|r\|_2^2$ on the control-point offset would instead assign the cost $\lambda_p\|q\|_2^2/[4t^2(1-t)^2]$ to the same $q$ when $t\in(0,1)$.
We therefore use $R_{\mathrm{disp}}(q)$ to assign a fixed displacement the same cost at different interior reference positions.

The endpoint segment provides an observable reference, while regularized displacement allows the evaluation center to adapt to the observed update.
To account for gradient variation, the shared two-state component evaluates all candidate images at the same pair of states around the adapted center and averages the two raw parameter gradients before matching.

\subsection{Shared Two-State Gradient Matching}
\label{subsec:shared_state_matching}

Let us define $v=w_T-w_0$ as the state direction obtained from the observed endpoints, without access to intermediate local states.
From Eq.~\eqref{eq:full_gradient_average}, a gradient at the adapted center $\hat{w}$ need not represent the average of full-data gradients across local states.
To approximate this average, we introduce the shared symmetric states $w_+=\hat{w}+hv$ and $w_-=\hat{w}-hv$, where $h\ge0$ controls their separation.
The surrogate evaluates $g_{\widehat{X}}$ at both states, with the same pair shared by all candidate images.
We average the two raw parameter gradients before computing a single cosine distance to the observed update:
\begin{subequations}
\label{eq:method-shared-matching}
\begin{align}
G_h(\widehat{X},\hat{w})
&=\tfrac12\bigl[g_{\widehat{X}}(w_+)+g_{\widehat{X}}(w_-)\bigr],
\label{eq:method-shared-gradient}\\*
\mathcal{L}_{\mathrm{cos},h}(\widehat{X},\hat{w})
&=1-\frac{\langle u,G_h(\widehat{X},\hat{w})\rangle}
 {\|u\|_2\|G_h(\widehat{X},\hat{w})\|_2}.
\label{eq:method-shared-cosine}
\end{align}
\end{subequations}
Both $u$ and $G_h$ are required to be nonzero for the cosine distance. Setting $h=0$ recovers the single-state matching signal, while $h>0$ samples gradients on both sides of the center.

Consider the two-state average for fixed $\widehat{X},\hat{w},v$, and assume that $g_{\widehat{X}}$ is twice continuously differentiable in a neighborhood of $\hat{w}$.
As $h\to0$, a Taylor expansion of $g_{\widehat{X}}$ at the center yields
\begin{equation}
\begin{aligned}
g_{\widehat{X}}(w_\pm)={}&g_{\widehat{X}}(\hat{w})
 \pm hDg_{\widehat{X}}(\hat{w})[v]\\
&+\frac{h^2}{2}D^2g_{\widehat{X}}(\hat{w})[v,v]+e_\pm(h),
\end{aligned}
\label{eq:method-node-expansion}
\end{equation}
where $Dg_{\widehat{X}}$ and $D^2g_{\widehat{X}}$ denote the first and second derivatives of the gradient map with respect to model parameters, respectively, and $\|e_\pm(h)\|_2=o(h^2)$.
The opposite first-order terms cancel when the two expansions are averaged.
Subtracting the gradient at the center then yields the second-order correction
\begin{equation}
\begin{aligned}
&G_h(\widehat{X},\hat{w})-g_{\widehat{X}}(\hat{w})\\
&\quad=\tfrac12\bigl[g_{\widehat{X}}(w_+)-2g_{\widehat{X}}(\hat{w})
                       +g_{\widehat{X}}(w_-)\bigr]\\
&\quad=\frac{h^2}{2}D^2g_{\widehat{X}}(\hat{w})[v,v]+\bar{e}(h),
\end{aligned}
\label{eq:method-shared-correction}
\end{equation}
where $\bar{e}(h)=[e_+(h)+e_-(h)]/2$ is the averaged remainder and satisfies $\|\bar{e}(h)\|_2=o(h^2)$.

The average in Eq.~\eqref{eq:method-shared-gradient} is computed directly from the two parameter gradients without explicitly forming the higher-order derivatives in Eq.~\eqref{eq:method-shared-correction}.
Each iteration therefore uses two model forward passes and two parameter-gradient evaluations, followed by differentiation of the joint reconstruction objective.
Moreover, we reuse these forward passes to average cross-entropy over the same states for $\Phi_h$ in Section~\ref{subsec:state_displacement}.
We optimize the width $h$ jointly with the images and center parameters and enforce $h\in[0,1/2]$ independently of $t$.
This constraint limits the state separation $2h\|v\|_2$ but does not restrict the two evaluation states to the endpoint segment.

A closer examination of Eq.~\eqref{eq:method-shared-correction} reveals how center displacement and two-state averaging affect the matching signal.
Notice that, for fixed candidate images and an endpoint reference, changing $q$ moves the evaluation center $\hat{w}$.
At a fixed center, the small-$h$ expansion describes the second-order directional correction controlled by $h$.
When the gradient map is affine on the segment joining the two evaluation states, $G_h(\widehat{X},\hat{w})=g_{\widehat{X}}(\hat{w})$, so averaging leaves the matching signal unchanged.
For a map that is nonlinear along this segment, averaging can retain a contribution that a single evaluation at the same $\hat{w}$ would omit.
This local interpretation does not imply that the learned $\hat{w},h$ recover the mean or spread of the hidden SGD states.
Moreover, the matching discrepancy can still include the mini-batch residual $\bar{\xi}$ identified in Section~\ref{subsec:problem_formulation}.
The two shared evaluations thus provide a compact model of local state variation, whose reconstruction benefit is assessed empirically.

\subsection{Concrete Formulation}
\label{subsec:method_optimization}

\textbf{Loss function.}
Our reconstruction objective consists of four terms, including shared two-state cosine matching, total variation (TV), label cross-entropy, and displacement regularization.
We employ the cosine distance in Eq.~\eqref{eq:method-shared-cosine} to compare the averaged raw parameter gradient $G_h(\widehat{X},\hat{w})$ with the observed update $u$.
This distance is invariant to the positive scaling relating the accumulated update $u$ to its average $\bar{u}$.
Moreover, TV penalizes differences between neighboring pixels using the average of the horizontal and vertical mean absolute differences, encouraging spatially smooth reconstructions.
The cross-entropy term averages the losses at $w_+$ and $w_-$ to encourage predictions consistent with the known labels $y$ at both states.
The displacement term applies a quadratic penalty $4\lambda_p\|q\|_2^2$ to the actual displacement of the evaluation center from its selected reference on the endpoint segment.
The complete reconstruction objective is formulated as follows:
\begin{equation}
\begin{aligned}
\mathcal{J}(\widehat{X},t,P_1,h)={}&\mathcal{L}_{\mathrm{cos},h}(\widehat{X},\hat{w})\\
&+\lambda_{\mathrm{TV}}\operatorname{TV}(\widehat{X})\\
&+\frac{\gamma}{2}\bigl[\ell(w_+;\widehat{X},y)+\ell(w_-;\widehat{X},y)\bigr]\\
&+4\lambda_p\|q\|_2^2.
\end{aligned}
\label{eq:method-objective}
\end{equation}
The weights $\lambda_{\mathrm{TV}},\gamma,\lambda_p$ balance the influence of TV, label cross-entropy, and displacement regularization, respectively.
For fixed $\widehat{X}$ and $h$, TV is constant with respect to the state parameters, so the remaining terms coincide with $H_{t,h}(q)$ analyzed in Section~\ref{subsec:state_displacement}.

\Needspace{5\baselineskip}
\textbf{Optimization.}
We employ separate Adam optimizers to jointly minimize the objective in Eq.~\eqref{eq:method-objective} over $\widehat{X},t,P_1,h$.
At each iteration, we compute all image and state derivatives at the same iterate before updating the four variable groups.
The normalized pixel box $\mathcal{X}$ has channelwise bounds $[-\mu_c/\sigma_c,(1-\mu_c)/\sigma_c]$, where $\mu_c$ and $\sigma_c>0$ are the known normalization mean and standard deviation.
To adhere to the domain constraints, we project $\widehat{X}$ onto $\mathcal{X}$, $t$ onto $[0,1]$, and $h$ onto $[0,1/2]$ after these updates.
Algorithm~\ref{alg:nlsme} summarizes the initialization and optimization steps, and returns the final iterate $\widehat{X}_K$ after inverse normalization.
\FloatBarrier

\Needspace{10\baselineskip}

\begin{table*}[!t] \centering
\caption{Reconstruction on FEMNIST and CIFAR-100 under different local-training lengths. Results are mean $\pm$ standard deviation over five runs. NL-SME is shaded.} \label{tab:femnist_cifar100_e20_e50_lsim}
\setlength{\tabcolsep}{2.6pt} \renewcommand{\arraystretch}{1.05}
\resizebox{\textwidth}{!}{%
\begin{tabular}{ll|cccccc|cccccc}
\toprule
\multirow{2}{*}{Dataset} & \multirow{2}{*}{Metric} & \multicolumn{6}{c|}{\makecell[c]{$E=20,N=50,B=10$ $(T=100)$}} & \multicolumn{6}{c}{\makecell[c]{$E=50,N=50,B=10$ $(T=250)$}} \\
\cmidrule(lr){3-8} \cmidrule(lr){9-14}
& & IG & SME & FEDLEAK & GI-NAS & DGGI & \cellcolor{gray!20}NL-SME & IG & SME & FEDLEAK & GI-NAS & DGGI & \cellcolor{gray!20}NL-SME \\
\midrule
\multirow{5}{*}{FEMNIST} & PSNR $\uparrow$ & 16.3543 $\pm$ 0.3081 & 21.3836 $\pm$ 0.2837 & 16.4234 $\pm$ 0.2143 & 16.3646 $\pm$ 0.3502 & 21.5867 $\pm$ 0.6073 & \cellcolor{gray!20}\textbf{24.1847 $\pm$ 0.3858} & 13.3076 $\pm$ 0.3898 & 20.3665 $\pm$ 0.2140 & 13.2911 $\pm$ 0.2735 & 13.1602 $\pm$ 0.4414 & 20.2827 $\pm$ 0.4597 & \cellcolor{gray!20}\textbf{22.6578 $\pm$ 0.2769} \\
& SSIM $\uparrow$ & 0.4955 $\pm$ 0.0231 & 0.7844 $\pm$ 0.0190 & 0.5028 $\pm$ 0.0118 & 0.5241 $\pm$ 0.0300 & 0.8014 $\pm$ 0.0220 & \cellcolor{gray!20}\textbf{0.8952 $\pm$ 0.0069} & 0.3728 $\pm$ 0.0318 & 0.7542 $\pm$ 0.0126 & 0.3635 $\pm$ 0.0280 & 0.4038 $\pm$ 0.0219 & 0.7505 $\pm$ 0.0219 & \cellcolor{gray!20}\textbf{0.8474 $\pm$ 0.0095} \\
& FSIM $\uparrow$ & 0.6873 $\pm$ 0.0141 & 0.8863 $\pm$ 0.0258 & 0.6906 $\pm$ 0.0108 & 0.7330 $\pm$ 0.0105 & 0.9072 $\pm$ 0.0178 & \cellcolor{gray!20}\textbf{0.9656 $\pm$ 0.0019} & 0.5995 $\pm$ 0.0148 & 0.8852 $\pm$ 0.0158 & 0.5977 $\pm$ 0.0126 & 0.6443 $\pm$ 0.0251 & 0.8889 $\pm$ 0.0221 & \cellcolor{gray!20}\textbf{0.9495 $\pm$ 0.0040} \\
& LPIPS $\downarrow$ & 0.3905 $\pm$ 0.0144 & 0.1314 $\pm$ 0.0372 & 0.3885 $\pm$ 0.0131 & 0.3289 $\pm$ 0.0109 & 0.1019 $\pm$ 0.0237 & \cellcolor{gray!20}\textbf{0.0377 $\pm$ 0.0038} & 0.4641 $\pm$ 0.0119 & 0.1283 $\pm$ 0.0203 & 0.4641 $\pm$ 0.0136 & 0.4001 $\pm$ 0.0213 & 0.1263 $\pm$ 0.0251 & \cellcolor{gray!20}\textbf{0.0540 $\pm$ 0.0037} \\
& $L_{\mathrm{sim}}$ $\downarrow$ & 0.1617 $\pm$ 0.0284 & 0.0347 $\pm$ 0.0031 & 0.1625 $\pm$ 0.0301 & 0.1858 $\pm$ 0.0454 & 0.0350 $\pm$ 0.0033 & \cellcolor{gray!20}0.0011 $\pm$ 0.0001 & 0.3102 $\pm$ 0.0378 & 0.0454 $\pm$ 0.0085 & 0.2997 $\pm$ 0.0364 & 0.3278 $\pm$ 0.0438 & 0.0486 $\pm$ 0.0060 & \cellcolor{gray!20}0.0025 $\pm$ 0.0004 \\
\midrule
\multirow{5}{*}{CIFAR-100} & PSNR $\uparrow$ & 14.8674 $\pm$ 0.3964 & 21.5318 $\pm$ 0.9798 & 14.3758 $\pm$ 0.3540 & 15.9058 $\pm$ 0.8671 & 21.2392 $\pm$ 0.5084 & \cellcolor{gray!20}\textbf{26.1211 $\pm$ 0.3956} & 12.0075 $\pm$ 0.2343 & 18.1818 $\pm$ 0.3870 & 11.7172 $\pm$ 0.2507 & 12.4511 $\pm$ 0.3878 & 18.4492 $\pm$ 0.3349 & \cellcolor{gray!20}\textbf{23.9413 $\pm$ 0.5881} \\
& SSIM $\uparrow$ & 0.3804 $\pm$ 0.0118 & 0.7278 $\pm$ 0.0303 & 0.3497 $\pm$ 0.0179 & 0.4399 $\pm$ 0.0546 & 0.7199 $\pm$ 0.0161 & \cellcolor{gray!20}\textbf{0.8794 $\pm$ 0.0112} & 0.2121 $\pm$ 0.0164 & 0.5754 $\pm$ 0.0197 & 0.1962 $\pm$ 0.0179 & 0.2353 $\pm$ 0.0177 & 0.5835 $\pm$ 0.0196 & \cellcolor{gray!20}\textbf{0.8019 $\pm$ 0.0214} \\
& FSIM $\uparrow$ & 0.7065 $\pm$ 0.0091 & 0.8553 $\pm$ 0.0139 & 0.6893 $\pm$ 0.0085 & 0.7213 $\pm$ 0.0179 & 0.8476 $\pm$ 0.0075 & \cellcolor{gray!20}\textbf{0.9333 $\pm$ 0.0074} & 0.6220 $\pm$ 0.0061 & 0.7696 $\pm$ 0.0114 & 0.6120 $\pm$ 0.0049 & 0.6279 $\pm$ 0.0065 & 0.7703 $\pm$ 0.0114 & \cellcolor{gray!20}\textbf{0.8862 $\pm$ 0.0106} \\
& LPIPS $\downarrow$ & 0.4800 $\pm$ 0.0120 & 0.2423 $\pm$ 0.0189 & 0.5004 $\pm$ 0.0077 & 0.4540 $\pm$ 0.0226 & 0.2528 $\pm$ 0.0135 & \cellcolor{gray!20}\textbf{0.1190 $\pm$ 0.0124} & 0.5729 $\pm$ 0.0034 & 0.3722 $\pm$ 0.0155 & 0.5822 $\pm$ 0.0040 & 0.5650 $\pm$ 0.0099 & 0.3755 $\pm$ 0.0153 & \cellcolor{gray!20}\textbf{0.2106 $\pm$ 0.0172} \\
& $L_{\mathrm{sim}}$ $\downarrow$ & 0.1385 $\pm$ 0.0114 & 0.0484 $\pm$ 0.0050 & 0.1313 $\pm$ 0.0118 & 0.1586 $\pm$ 0.0099 & 0.0504 $\pm$ 0.0030 & \cellcolor{gray!20}0.0035 $\pm$ 0.0006 & 0.2681 $\pm$ 0.0160 & 0.0610 $\pm$ 0.0060 & 0.2543 $\pm$ 0.0179 & 0.2914 $\pm$ 0.0200 & 0.0610 $\pm$ 0.0045 & \cellcolor{gray!20}0.0053 $\pm$ 0.0002 \\
\bottomrule
\end{tabular}%
}
\end{table*}

\begin{figure*}[!t] \centering
\begin{minipage}[t]{0.485\textwidth} \centering \includegraphics[width=\linewidth]{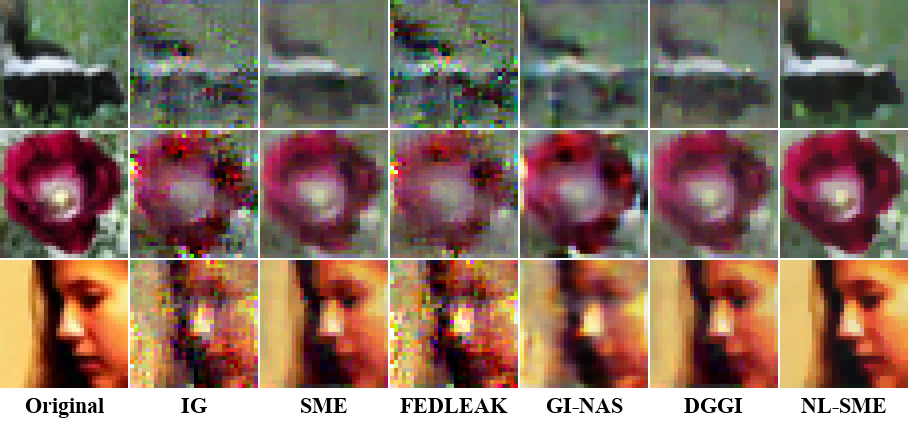} \vspace{1mm} {\footnotesize \textbf{(a)} CIFAR-100\par} \end{minipage}
\hfill
\begin{minipage}[t]{0.485\textwidth} \centering \includegraphics[width=\linewidth]{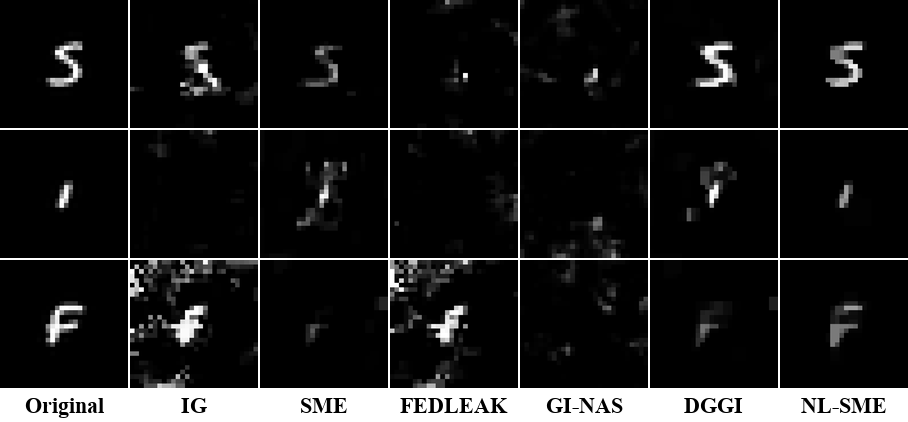} \vspace{1mm} {\footnotesize \textbf{(b)} FEMNIST\par} \end{minipage}
\caption{Image reconstruction under the default setting. Each row shows the same original image across methods.} \label{fig:qualitative_cifar100_femnist}
\end{figure*}

\section{Experiments}
\label{sec:experiments}

\subsection{Experimental Setup}
\label{subsec:experimental_setup}

\indent\textbf{Evaluation Settings.}
We test our method on FEMNIST~\cite{caldas2018leaf} and CIFAR-100~\cite{krizhevsky2009learning}.
We construct multi-step FL settings by specifying the number of local epochs, local data size, and batch size.
Following the model settings in~\cite{dimitrov2022data}, we adopt CNNmnist and CNNcifar as the default global models for FEMNIST and CIFAR-100, respectively.
We evaluate local updates from randomly initialized victim models.
Appendix~\ref{appendix:additional_results} reports supplementary results with MLP~\cite{zhu2023surrogate} and ViT~\cite{dosovitskiy2021image}, as well as a Tiny-ImageNet~\cite{le2015tiny} sanity check using a three-layer MLP.

\indent\textbf{Baselines.}
We compare NL-SME with five gradient inversion methods.
IG~\cite{geiping2020inverting} uses a cosine objective for gradient matching at $w_0$.
SME~\cite{zhu2023surrogate} learns an interpolation between $w_0$ and $w_T$ for gradient matching.
FEDLEAK~\cite{fan2025boosting} combines partial-gradient $\ell_1$ and cosine matching at $w_0$ with gradient regularization.
GI-NAS~\cite{yu2025ginas} searches for a neural architecture before reconstructing images through gradient matching at $w_0$.
We adapt DGGI~\cite{wu2025dggi} to combine diffusion-based generation and alignment with gradient matching at a learned interpolation between the model endpoints.
These implementations match the observed update $u=w_0-w_T$ without unrolling local SGD during reconstruction, with cosine terms using only its direction.
In our comparisons, all methods, including FEDLEAK, use ground-truth labels that remain fixed during reconstruction.
For the supplementary evaluation on trained networks, we replace
DGGI with DLFA~\cite{dimitrov2022data}, which differentiates through
simulated local SGD updates.

\indent\textbf{Quantitative Metrics.}
We report four image-quality metrics: Peak Signal-to-Noise Ratio (PSNR) $\uparrow$, Structural Similarity Index Measure (SSIM) $\uparrow$, Feature Similarity Index Measure (FSIM) $\uparrow$, and Learned Perceptual Image Patch Similarity (LPIPS) $\downarrow$.
Here, $\uparrow$ indicates that higher values are better, while $\downarrow$ indicates that lower values are better.
For each method and run, we use linear sum assignment to obtain a one-to-one pairing that minimizes total pixel-wise MSE without label constraints.
The same pairing is used to compute PSNR, SSIM, FSIM, and LPIPS.
We additionally report the method-specific similarity loss \(L_{\mathrm{sim}}\) $\downarrow$ to assess fit to each attack's matching objective.
These losses are not directly comparable across methods.

\indent\textbf{Implementation Details.}
In the main benchmark comparisons, each method runs for $1000$ iterations in the final reconstruction stage.
NL-SME evaluates two parameter gradients per iteration, whereas IG and SME evaluate one, so equal iteration counts do not imply equal computational budgets.
For NL-SME, we adopt separate Adam optimizers for the candidate images and state parameters.
We use cosine distance to match the average of the raw parameter gradients at the two shared model states with the observed update.
The number of local steps is computed as $T=E\lceil N/B\rceil$, where $E$, $N$, and $B$ denote the number of local epochs, local data size, and batch size, respectively.
The main benchmark and clipping comparisons use the same private samples and model endpoints across methods.
All methods are run on an NVIDIA RTX 5070 Ti GPU.
Additional experimental details are provided in Appendix~\ref{appendix:experimental_details}.

\subsection{Attack in Benchmark Datasets}
\label{subsec:benchmark_attack}

Table~\ref{tab:femnist_cifar100_e20_e50_lsim} presents the average PSNR of reconstructed images using six different attack methods, evaluated on FEMNIST and CIFAR-100 with different numbers of local steps.
Throughout these settings, NL-SME outperforms the other attack methods.
For instance, on CIFAR-100 with $250$ local steps, NL-SME achieves a PSNR of $23.94$ dB, while DGGI and SME achieve $18.45$ and $18.18$ dB, respectively.
Figure~\ref{fig:qualitative_cifar100_femnist} shows reconstructed images for both datasets.

\begin{figure*}[!t] \centering
\begin{minipage}[t]{0.315\textwidth} \centering \includegraphics[width=\linewidth]{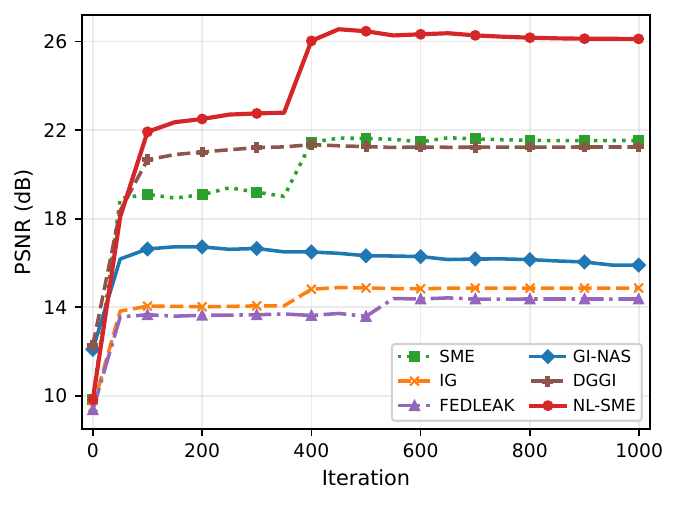} \vspace{1mm} \refstepcounter{figure} \label{fig:cifar100_iteration_psnr} \parbox{\linewidth}{\footnotesize\textbf{Figure~\thefigure:} Reconstruction quality over attack iterations.} \end{minipage}
\hfill
\begin{minipage}[t]{0.315\textwidth} \centering \includegraphics[width=\linewidth]{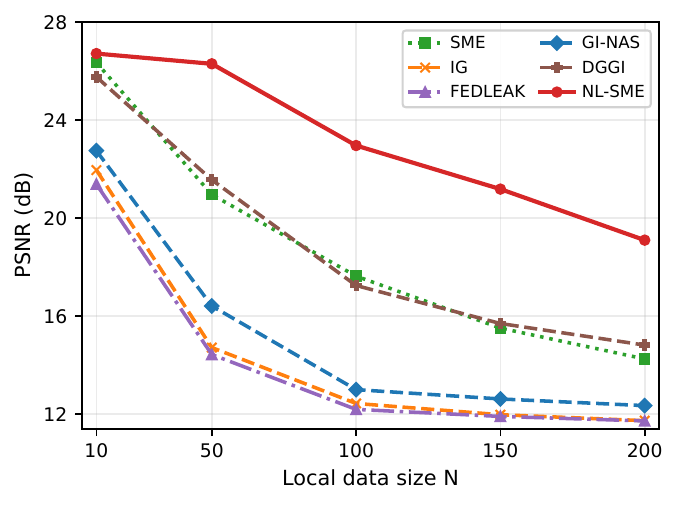} \vspace{1mm} \refstepcounter{figure} \label{fig:cifar100_varying_n_psnr} \parbox{\linewidth}{\footnotesize\textbf{Figure~\thefigure:} Effect of local data size on reconstruction.} \end{minipage}
\hfill
\begin{minipage}[t]{0.315\textwidth} \centering \includegraphics[width=\linewidth]{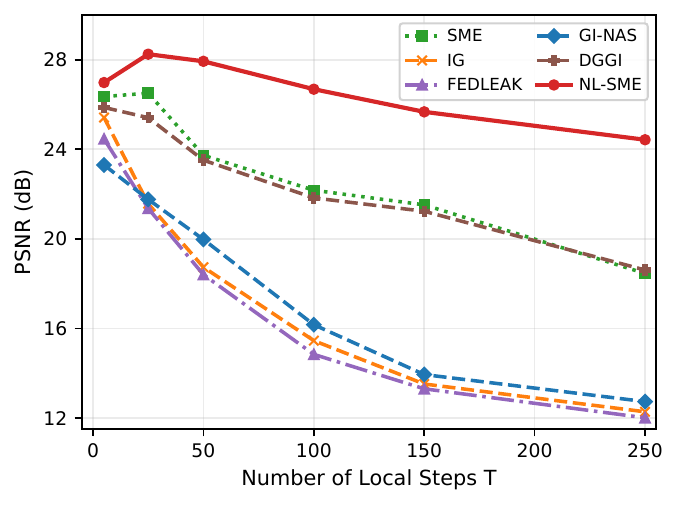} \vspace{1mm} \refstepcounter{figure} \label{fig:cifar100_varying_e_local_steps_psnr} \parbox{\linewidth}{\footnotesize\textbf{Figure~\thefigure:} Effect of local training length on reconstruction.} \end{minipage}
\end{figure*}

\begin{table}[!t] \centering
\caption{Reconstruction from fused client updates on CIFAR-100.} \label{tab:fused_update_cifar100}
\scriptsize \setlength{\tabcolsep}{3.2pt} \renewcommand{\arraystretch}{1.05}
\resizebox{\columnwidth}{!}{%
\begin{tabular}{@{}lcccc@{}}
\toprule
\textbf{Method} & \textbf{PSNR}$\uparrow$ & \textbf{SSIM}$\uparrow$ & \textbf{LPIPS}$\downarrow$ & \textbf{FSIM}$\uparrow$ \\
\midrule
IG & 13.3497 $\pm$ 0.2595 & 0.2948 $\pm$ 0.0326 & 0.5305 $\pm$ 0.0134 & 0.6591 $\pm$ 0.0213 \\
SME & 17.8687 $\pm$ 0.9033 & 0.5654 $\pm$ 0.0616 & 0.3680 $\pm$ 0.0392 & 0.7817 $\pm$ 0.0288 \\
FEDLEAK & 13.0144 $\pm$ 0.1777 & 0.2713 $\pm$ 0.0254 & 0.5463 $\pm$ 0.0118 & 0.6460 $\pm$ 0.0201 \\
GI-NAS & 14.1153 $\pm$ 0.5636 & 0.3354 $\pm$ 0.0303 & 0.5160 $\pm$ 0.0369 & 0.6672 $\pm$ 0.0198 \\
DGGI & 17.9477 $\pm$ 0.7006 & 0.5690 $\pm$ 0.0545 & 0.3619 $\pm$ 0.0361 & 0.7802 $\pm$ 0.0250 \\
\cellcolor{gray!20}NL-SME & \cellcolor{gray!20}\textbf{25.0813 $\pm$ 0.5476} & \cellcolor{gray!20}\textbf{0.8488 $\pm$ 0.0202} & \cellcolor{gray!20}\textbf{0.1495 $\pm$ 0.0202} & \cellcolor{gray!20}\textbf{0.9169 $\pm$ 0.0117} \\
\bottomrule
\end{tabular}%
}
\end{table}

\begin{table*}[!t] \centering
\caption{Reconstruction under gradient sparsification and per-example gradient clipping on FEMNIST.} \label{tab:defense_comparison}
\scriptsize \setlength{\tabcolsep}{1.8pt} \renewcommand{\arraystretch}{1.03}
\resizebox{\textwidth}{!}{%
\begin{tabular}{@{}lccccccc@{\quad}ccccccc@{}}
\toprule
\textbf{Metric} & \textbf{Defense} & \textbf{IG} & \textbf{SME} & \textbf{FEDLEAK} & \textbf{GI-NAS} & \textbf{DGGI} & \cellcolor{gray!20}\textbf{NL-SME} & \textbf{Defense} & \textbf{IG} & \textbf{SME} & \textbf{FEDLEAK} & \textbf{GI-NAS} & \textbf{DGGI} & \cellcolor{gray!20}\textbf{NL-SME} \\
\midrule
PSNR $\uparrow$ & \multirow{4}{*}{\shortstack[c]{Gradient\\Sparsification}} & 16.7696 $\pm$ 0.2406 & 20.7015 $\pm$ 0.5943 & 16.7492 $\pm$ 0.3088 & 16.7822 $\pm$ 0.3725 & 20.8734 $\pm$ 0.4949 & \cellcolor{gray!20}\textbf{23.0246 $\pm$ 0.1204} & \multirow{4}{*}{\shortstack[c]{Gradient\\Clipping}} & 16.5566 $\pm$ 0.3177 & 21.2822 $\pm$ 0.3986 & 16.4095 $\pm$ 0.3325 & 16.6777 $\pm$ 0.5000 & 21.2873 $\pm$ 0.3756 & \cellcolor{gray!20}\textbf{24.2831 $\pm$ 0.2414} \\
SSIM $\uparrow$ &  & 0.5240 $\pm$ 0.0172 & 0.7604 $\pm$ 0.0360 & 0.5227 $\pm$ 0.0219 & 0.5520 $\pm$ 0.0276 & 0.7753 $\pm$ 0.0291 & \cellcolor{gray!20}\textbf{0.8512 $\pm$ 0.0103} &  & 0.5006 $\pm$ 0.0243 & 0.7803 $\pm$ 0.0250 & 0.4976 $\pm$ 0.0205 & 0.5373 $\pm$ 0.0263 & 0.7781 $\pm$ 0.0236 & \cellcolor{gray!20}\textbf{0.8946 $\pm$ 0.0056} \\
FSIM $\uparrow$ &  & 0.6982 $\pm$ 0.0086 & 0.8605 $\pm$ 0.0227 & 0.6984 $\pm$ 0.0070 & 0.7317 $\pm$ 0.0077 & 0.8764 $\pm$ 0.0177 & \cellcolor{gray!20}\textbf{0.9212 $\pm$ 0.0081} &  & 0.6927 $\pm$ 0.0145 & 0.8836 $\pm$ 0.0255 & 0.6914 $\pm$ 0.0158 & 0.7313 $\pm$ 0.0094 & 0.8977 $\pm$ 0.0149 & \cellcolor{gray!20}\textbf{0.9658 $\pm$ 0.0015} \\
LPIPS $\downarrow$ &  & 0.3698 $\pm$ 0.0103 & 0.1647 $\pm$ 0.0315 & 0.3674 $\pm$ 0.0089 & 0.3179 $\pm$ 0.0225 & 0.1412 $\pm$ 0.0256 & \cellcolor{gray!20}\textbf{0.0839 $\pm$ 0.0058} &  & 0.3846 $\pm$ 0.0150 & 0.1316 $\pm$ 0.0346 & 0.3873 $\pm$ 0.0184 & 0.3277 $\pm$ 0.0166 & 0.1123 $\pm$ 0.0212 & \cellcolor{gray!20}\textbf{0.0374 $\pm$ 0.0007} \\
\bottomrule
\end{tabular}%
}
\end{table*}

\begin{table}[!t] \centering
\caption{NL-SME reconstruction under per-example clipping and Gaussian noise on FEMNIST.} \label{tab:dpsgd_defense}
\scriptsize \setlength{\tabcolsep}{4.2pt} \renewcommand{\arraystretch}{1.08}
\resizebox{\columnwidth}{!}{%
\begin{tabular}{cccccc}
\toprule
$\sigma$ & $4.799\!\times\!10^{-5}$ & $4.799\!\times\!10^{-4}$ & $0.004799$ & $0.04799$ & $0.4799$ \\
\midrule
PSNR $\uparrow$ & 23.9329 $\pm$ 0.5859 & 23.8165 $\pm$ 0.8245 & 23.4528 $\pm$ 0.9668 & 18.9840 $\pm$ 0.3279 & 15.6351 $\pm$ 0.1923 \\
\bottomrule
\end{tabular}%
}
\end{table}

\subsection{Attack in Diverse Settings}
\label{subsec:diverse_settings}

In this subsection, we examine how reconstruction quality changes during optimization and how local data size affects attack performance.
We will subsequently increase the number of local steps to create more challenging attack configurations.
Finally, we evaluate reconstruction from mixtures of client updates.

\textbf{Convergence behavior.}
Figure~\ref{fig:cifar100_iteration_psnr} shows the convergence curves under the default CIFAR-100 setting, where we observe that NL-SME achieves a PSNR of $26.03$ dB after approximately $400$ iterations.
Beyond this point, its PSNR shows only small changes.

\textbf{Evaluation over varying local data sizes.}
We use CIFAR-100 and vary the local data size from $10$ to $200$, with the number of local epochs and the batch size fixed.
As shown in Figure~\ref{fig:cifar100_varying_n_psnr}, NL-SME maintains higher PSNR than the other attack methods across all tested data sizes.
We see that its PSNR decreases as the local dataset grows, while the number of local updates also increases.

\textbf{Evaluation over varying local steps.}
Figure~\ref{fig:cifar100_varying_e_local_steps_psnr} presents the performance of six attack methods on CIFAR-100 when clients execute $5$ to $250$ local steps, with the local data size and batch size fixed.
A key observation is that NL-SME achieves higher PSNR than the other attack methods.
Moreover, its PSNR increases to its highest value at $25$ local steps and then declines as the number of local steps increases.

\textbf{Attack under fused update observations.}
Under the default CIFAR-100 setting, we mix the target-client update with the average update from four background clients, using weights of $0.75$ and $0.25$, respectively.
The results in Table~\ref{tab:fused_update_cifar100} show that NL-SME outperforms the other attack methods across all four metrics.
Specifically, NL-SME achieves a PSNR of $25.08$ dB, while SME achieves $17.87$ dB.
This experiment does not evaluate a secure-aggregation protocol.

\subsection{Attack under Representative Defense Strategies}
\label{subsec:defense_evaluation}

We now evaluate NL-SME against perturbation-based defenses~\cite{yu2025ginas,fan2025boosting} on FEMNIST.
We employ gradient sparsification, per-example gradient clipping, and per-example clipping with Gaussian noise in clients' local training.

\textbf{Attacks under sparsification and clipping.}
We consider gradient sparsification~\cite{aji2017sparse} with a pruning rate of \(90\%\) and per-example gradient clipping with a clipping bound of $4$~\cite{wei2021gradient}.
At each local step, sparsification prunes gradient elements within each parameter tensor, while clipping is applied to each sample gradient before averaging.
NL-SME uses global matching in both settings, without defense-specific masking or weighting.
As shown in Table~\ref{tab:defense_comparison}, NL-SME outperforms the other attack methods across all four metrics under both defenses.
For instance, under gradient clipping, NL-SME achieves a PSNR of $24.28$ dB, while DGGI achieves $21.29$ dB.

\textbf{Attacks under clipping and Gaussian noise.}
We employ per-example clipping with \(C=1\) and Gaussian noise at each local training step.
We add independent Gaussian noise with standard deviation \(\sigma C/B\) to each coordinate of the averaged clipped gradient.
For the two largest noise multipliers \(\sigma\), we use separate attack settings, fixed across seeds, as detailed in Appendix~\ref{appendix:experimental_details}.
As shown in Table~\ref{tab:dpsgd_defense}, PSNR changes little across the three smallest noise multipliers and is lower at the two largest.
At the largest noise multiplier, NL-SME achieves a PSNR of $15.64$ dB.
The noise multipliers specify noise strength, while composed privacy budgets are not reported because multi-step privacy accounting is not performed.
Model utility is not compared across these noise settings.

\begin{table}[!t] \centering
\caption{Ablation study of the two NL-SME components.} \label{tab:cifar100_trajectory_calibration_ablation}
\scriptsize \setlength{\tabcolsep}{3.0pt} \renewcommand{\arraystretch}{1.08}
\resizebox{\columnwidth}{!}{%
\begin{tabular}{ccccccc}
\toprule
\makecell[c]{Displacement\\$q$} & \makecell[c]{Two-state\\average} & PSNR $\uparrow$ & SSIM $\uparrow$ & LPIPS $\downarrow$ & FSIM $\uparrow$ & $L_{\mathrm{sim}}$ \\
\midrule
$\times$ & $\times$ & 21.5425 $\pm$ 0.8205 & 0.7243 $\pm$ 0.0280 & 0.2449 $\pm$ 0.0195 & 0.8548 $\pm$ 0.0116 & 0.0482 $\pm$ 0.0031 \\
$\times$ & $\checkmark$ & 24.1715 $\pm$ 0.6694 & 0.8184 $\pm$ 0.0108 & 0.1697 $\pm$ 0.0120 & 0.9045 $\pm$ 0.0066 & 0.0251 $\pm$ 0.0034 \\
$\checkmark$ & $\times$ & 25.0331 $\pm$ 0.3962 & 0.8644 $\pm$ 0.0080 & 0.1254 $\pm$ 0.0135 & 0.9262 $\pm$ 0.0066 & 0.0045 $\pm$ 0.0004 \\
$\checkmark$ & $\checkmark$ & \textbf{26.1211 $\pm$ 0.3956} & \textbf{0.8794 $\pm$ 0.0112} & \textbf{0.1190 $\pm$ 0.0124} & \textbf{0.9333 $\pm$ 0.0074} & \textbf{0.0035 $\pm$ 0.0006} \\
\bottomrule
\end{tabular}%
}
\end{table}

\begin{table}[!t] \centering
\caption{Sensitivity to the displacement regularization weight $\lambda_p$.} \label{tab:cifar100_lambda_p_sensitivity}
\scriptsize \setlength{\tabcolsep}{3.0pt} \renewcommand{\arraystretch}{1.08}
\resizebox{\columnwidth}{!}{%
\begin{tabular}{cccccc}
\toprule
$\lambda_p$ & 0 & $3\times10^{-5}$ & $3\times10^{-4}$ & $3\times10^{-3}$ & $3\times10^{-2}$ \\
\midrule
PSNR $\uparrow$ & 20.8280 $\pm$ 0.4671 & 21.7339 $\pm$ 0.4692 & 25.2623 $\pm$ 0.3994 & \textbf{26.1211 $\pm$ 0.3956} & 25.4867 $\pm$ 0.8706 \\
LPIPS $\downarrow$ & 0.2662 $\pm$ 0.0226 & 0.2362 $\pm$ 0.0119 & 0.1413 $\pm$ 0.0106 & \textbf{0.1190 $\pm$ 0.0124} & 0.1314 $\pm$ 0.0162 \\
\bottomrule
\end{tabular}%
}
\end{table}

\begin{table}[!t] \centering
\caption{Evaluation width and gradient budget with regularized displacement. Time covers all attack stages.} \label{tab:cifar100_width_budget}
\footnotesize \setlength{\tabcolsep}{3pt} \renewcommand{\arraystretch}{1.25}
\resizebox{\columnwidth}{!}{%
\begin{tabular*}{\textwidth}{@{\extracolsep{\fill}}lccccccc@{}}
\toprule
\multirow{2}{*}{Evaluation scheme} & \multicolumn{2}{c}{Computational budget} & \multicolumn{4}{c}{Reconstruction quality} & \multirow{2}{*}{Time (s)} \\
\cmidrule(lr){2-3}\cmidrule(lr){4-7}
& Iterations $K$ & \makecell{Gradient\\evaluations} & PSNR $\uparrow$ & SSIM $\uparrow$ & FSIM $\uparrow$ & LPIPS $\downarrow$ & \\
\midrule
Single state & 1000 & 1000 & 25.0331 $\pm$ 0.3962 & 0.8644 $\pm$ 0.0080 & 0.9262 $\pm$ 0.0066 & 0.1254 $\pm$ 0.0135 & 19.87 $\pm$ 1.96 \\
Single state & 2000 & 2000 & 24.9591 $\pm$ 0.4323 & 0.8615 $\pm$ 0.0091 & 0.9253 $\pm$ 0.0067 & 0.1253 $\pm$ 0.0122 & 38.55 $\pm$ 2.15 \\
\addlinespace[3pt]
Two states, fixed $h$ & 1000 & 2000 & 26.0078 $\pm$ 0.4228 & 0.8795 $\pm$ 0.0157 & 0.9324 $\pm$ 0.0084 & 0.1192 $\pm$ 0.0164 & 25.49 $\pm$ 0.18 \\
Two states, learned $h$ & 1000 & 2000 & 26.1211 $\pm$ 0.3956 & 0.8794 $\pm$ 0.0112 & 0.9333 $\pm$ 0.0074 & 0.1190 $\pm$ 0.0124 & 25.35 $\pm$ 0.36 \\
\bottomrule
\end{tabular*}%
}
\end{table}

\begin{table}[!t] \centering
\caption{Comparison with SME at $2000$ matching parameter-gradient evaluations on CIFAR-100.} \label{tab:cifar100_external_sme_budget}
\footnotesize \setlength{\tabcolsep}{3pt} \renewcommand{\arraystretch}{1.25}
\resizebox{\columnwidth}{!}{%
\begin{tabular*}{\textwidth}{@{\extracolsep{\fill}}lcccccc@{}}
\toprule
\multirow{2}{*}{Method} & \multicolumn{2}{c}{Computational budget} & \multicolumn{4}{c}{Reconstruction quality} \\
\cmidrule(lr){2-3}\cmidrule(lr){4-7}
& Iterations $K$ & \makecell{Gradient\\evaluations} & PSNR $\uparrow$ & SSIM $\uparrow$ & FSIM $\uparrow$ & LPIPS $\downarrow$ \\
\midrule
SME & 2000 & 2000 & 21.4734 $\pm$ 0.8759 & 0.7253 $\pm$ 0.0294 & 0.8537 $\pm$ 0.0129 & 0.2418 $\pm$ 0.0182 \\
\rowcolor{gray!20}
NL-SME (two shared states) & 1000 & 2000 & \textbf{26.1211 $\pm$ 0.3956} & \textbf{0.8794 $\pm$ 0.0112} & \textbf{0.9333 $\pm$ 0.0074} & \textbf{0.1190 $\pm$ 0.0124} \\
\bottomrule
\end{tabular*}%
}
\end{table}

\begin{table}[!t] \centering
\caption{The time and memory requirements of different attack methods on CIFAR-100.} \label{tab:computational_cost}
\scriptsize \setlength{\tabcolsep}{4pt} \resizebox{\columnwidth}{!}{%
\begin{tabular}{lccccc>{\columncolor{gray!20}}c}
\toprule
Metric & IG & SME & FEDLEAK & GI-NAS & DGGI & NL-SME \\
\midrule
Time (s)         & 15.66  & 16.52  & 47.24   & 59.28  & 80.93  & 25.35  \\
Peak memory (MB) & 203.31 & 203.31 & 1084.09 & 304.54 & 820.90 & 338.60 \\
\bottomrule
\end{tabular}}
\end{table}

\subsection{Further Analysis}
\label{subsec:further_analysis}

\textbf{Ablation study.}
We assess the effects of regularized state displacement and shared two-state gradient matching by disabling each component.
Setting \(q=0\) restricts the evaluation center to the endpoint segment, while \(h=0\) reduces the two-state gradient average to the gradient at the center.
As reported in Table~\ref{tab:cifar100_trajectory_calibration_ablation}, regularized displacement alone increases PSNR from $21.54$ to $25.03$ dB.

Shared two-state gradient matching also improves reconstruction without center displacement, while combining both components gives the best results across the four image-quality metrics.
Adding two-state gradient averaging to the displaced center further increases PSNR by $1.09$ dB, showing the benefit of combining both components in this setting.

\textbf{The impact of the regularization weight.}
Table~\ref{tab:cifar100_lambda_p_sensitivity} presents the impact of the displacement regularization weight \(\lambda_p\) on reconstruction quality.
As the weight increases, reconstruction quality first improves and then declines.
All nonzero weights in the table improve PSNR and LPIPS over the unregularized case, showing the benefit of regularizing the evaluation center's displacement.

\textbf{Evaluation width and gradient budget.}
We further assess the performance of NL-SME variants with fixed
and learned evaluation widths under regularized displacement.
As reported in Table~\ref{tab:cifar100_width_budget}, both
two-state variants achieve higher PSNR and lower LPIPS than
the single-state variant with the same number of
parameter-gradient evaluations.
Learning the width provides only small further improvements
in PSNR, FSIM, and LPIPS over a fixed width, while SSIM
remains similar.

\textbf{Evaluation under equal gradient budgets.}
Table~\ref{tab:cifar100_external_sme_budget} presents the
performance of NL-SME and the original SME under the default
CIFAR-100 setting.
Both methods use the same private samples, ground-truth labels,
initial candidate images, and endpoint models.
SME runs for $2000$ iterations and NL-SME for $1000$ iterations
with two shared states, giving each method $2000$ matching
parameter-gradient evaluations.
NL-SME outperforms SME across all four metrics, with a mean
PSNR improvement of $4.65$ dB.
These evaluation counts do not imply equal running time
or total computation.

\textbf{Computational cost.}
Table~\ref{tab:computational_cost} reports the mean running time
and peak allocated GPU memory over five runs.
The measurements include all attack stages, including architecture
search, diffusion prior generation, and alignment.
Input preparation, client training, output saving, and final
evaluation are excluded.
NL-SME requires $25.35$ seconds per attack, while SME requires
$16.52$ seconds.
FEDLEAK, GI-NAS, and DGGI require more time.
NL-SME also has lower peak allocated GPU memory than FEDLEAK
and DGGI.
While NL-SME incurs higher computational costs than IG and SME,
it achieves better reconstruction quality in this setting.

\FloatBarrier

\section{Conclusion}
\label{sec:conclusion}

In this paper, we present NL-SME, a regularized two-state gradient surrogate for reconstructing private data from multi-step client updates.
We first analyze the single-state approximation of a normalized multi-step update, accounting for both state variation and the mini-batch residual.
We then allow the evaluation center to move beyond the endpoint segment while penalizing its actual displacement.
All candidate images share two symmetric states around this center, and the two raw parameter gradients are averaged before a single cosine comparison with the observed update.
We jointly optimize the candidate images and state parameters without replaying the client's local training.
Our results show that combining regularized center displacement with shared two-state gradient averaging improves reconstruction from multi-step updates.

\textbf{Limitations and Future Directions.}
Although our analysis shows that center displacement can improve the regularized objective under the stated conditions, it does not guarantee unique image recovery or convergence of joint optimization.
The attack assumes known labels and observable individual client updates, excluding settings where secure aggregation hides these updates.
Reconstruction results do not establish a new privacy guarantee.
In the future, we hope to extend the analysis to joint optimization and evaluate larger models.

\section*{CRediT authorship contribution statement}

\textbf{Li Xia:} Writing -- review \& editing, Writing -- original draft, Visualization, Validation, Software, Methodology, Investigation, Formal analysis, Data curation, Conceptualization.
\textbf{Jing Yu:} Supervision, Project administration, Methodology, Investigation.
\textbf{Zheng Liu:} Resources, Formal analysis, Project administration.
\textbf{Sili Huang:} Supervision, Project administration, Conceptualization.
\textbf{Wei Tang:} Supervision, Methodology, Conceptualization.
\textbf{Xuan Liu:} Supervision, Project administration, Methodology, Investigation.

\section*{Declaration of competing interest}

The authors declare that they have no known competing financial interests or personal relationships that could have appeared to influence the work reported in this paper.

\section*{Data availability}
Data will be made available on request.

\appendix
\section{Experimental Details}
\label{appendix:experimental_details}

For the main CNN experiments, we use $N=50$ local samples,
a batch size of $B=10$, $E=20$ local epochs, and a local SGD
learning rate of $0.004$ unless otherwise specified.
For NL-SME, we use an initial image learning rate of $1$
and set $\lambda_{\mathrm{TV}}=0.01$ and $\lambda_p=0.003$.
The initial learning rates for $t$ and $h$ are both $10^{-3}$,
while that for $P_1$ is $10^{-3}$ on CIFAR-100 and
$3\times10^{-3}$ on FEMNIST.
We set $\gamma=0.0045$ on CIFAR-100 and $\gamma=0.01$ on FEMNIST.
We initialize the width at $h_0=1/\sqrt{12}$ and use the
same value for the fixed-width control.
All CIFAR-100 ablation variants use the same cross-entropy
weight, whereas SME has no additional cross-entropy regularizer.
Hyperparameters remain fixed across runs within each setting.
For the main benchmarks and external SME comparison, we report
means and sample standard deviations over five runs.
We draw new private samples for each CIFAR-100 run and use
the same client samples across FEMNIST runs.

For the Gaussian-noise experiments with $\sigma=0.04799$
and $0.4799$, we use TV weights of $0.05$ and $0.25$ and
initial image learning rates of $0.3$ and $0.05$, respectively.
In both settings, the initial learning rate of $P_1$ is
$10^{-3}$.
We use the same gradient-matching loss for clean and
defended updates.

\section{Proof Details}
\label{appendix:supporting_theory}
\label{appendix:state_displacement}
\label{subsec:update_weights}
\label{appendix:legacy_state_regularization}

\begin{proof}[\normalfont\bfseries Proof of Theorem~\ref{thm:representative_state_mismatch}]
From Eq.~\eqref{eq:full_gradient_average}, the state residual is
\begin{equation}
\delta_{\mathrm{state}}(\widetilde w)
=\frac1T\sum_{\tau=0}^{T-1}
 [g_X(\widetilde w)-g_X(w_\tau)]-\bar\xi.
\label{eq:app_state_error_decomposition}
\end{equation}
Taking norms and applying the triangle inequality gives
\begin{equation}
\|\delta_{\mathrm{state}}(\widetilde w)\|_2
\le\frac1T\sum_{\tau=0}^{T-1}
 \|g_X(\widetilde w)-g_X(w_\tau)\|_2+\|\bar\xi\|_2.
\label{eq:app_state_error_triangle}
\end{equation}
Using the $L$-Lipschitz condition on the stated region and the definition of $\varepsilon_{\mathrm{state}}$, we obtain
\begin{equation}
\|\delta_{\mathrm{state}}(\widetilde w)\|_2
\le L\varepsilon_{\mathrm{state}}(\widetilde w)+\|\bar\xi\|_2.
\label{eq:app_representative_state_mismatch_bound}
\end{equation}
This completes the proof.
\end{proof}

To derive Eq.~\eqref{eq:single_state_direction}, let $a=g_X(\widetilde w)\ne0$ and $b=\bar u\ne0$.
Expanding $\|a-b\|_2^2$ yields
\begin{equation}
1-\cos(a,b)
=\frac{\|a-b\|_2^2-(\|a\|_2-\|b\|_2)^2}
 {2\|a\|_2\|b\|_2}.
\label{eq:app_cosine_identity}
\end{equation}
Since $u=\eta_{\mathrm{tr}}Tb$ with $\eta_{\mathrm{tr}}T>0$, $\cos(a,u)=\cos(a,b)$.
Dropping the nonnegative squared norm difference and using Eq.~\eqref{eq:app_representative_state_mismatch_bound} gives Eq.~\eqref{eq:single_state_direction}.
This single-state bound neither guarantees image recovery nor directly bounds the averaged-gradient discrepancy.

\begin{proof}[\normalfont\bfseries Proof of Theorem~\ref{thm:regularized_center_adaptation}]
Fix $\widehat X,y$ and $h\in[0,1/2]$, with $\lambda_p>0$ and $v=w_T-w_0\ne0$.
Let $\mathcal U$ be the convex neighborhood in the theorem, $d_\Phi=\nabla\Phi_h(w_{\mathrm{seg}})\ne0$, and $\kappa=L_\Phi+8\lambda_p>0$.
The cosine inputs are nonzero on the endpoint segment and $\mathcal U$.
Since $t_{\mathrm{seg}}\in(0,1)$ minimizes the objective along the segment,
\begin{equation}
\left.\frac{d}{dt}\Phi_h(w_{\mathrm{lin}}(t))\right|_{t=t_{\mathrm{seg}}}
=\langle d_\Phi,v\rangle=0.
\label{eq:app_segment_stationarity}
\end{equation}
For $w_{\mathrm{seg}}+q\in\mathcal U$, the descent lemma and displacement penalty give
\begin{equation}
H_{t_{\mathrm{seg}},h}(q)-H_{t_{\mathrm{seg}},h}(0)
\le\langle d_\Phi,q\rangle+\frac{\kappa}{2}\|q\|_2^2.
\label{eq:app_center_descent_bound}
\end{equation}
Choose $q_\alpha=-\alpha d_\Phi$ with $0<\alpha\le\kappa^{-1}$ small enough that $w_{\mathrm{seg}}+q_\alpha\in\mathcal U$.
Substituting into Eq.~\eqref{eq:app_center_descent_bound} yields
\begin{equation}
H_{t_{\mathrm{seg}},h}(q_\alpha)-H_{t_{\mathrm{seg}},h}(0)
\le-\alpha\bigl(1-\tfrac{\kappa\alpha}{2}\bigr)\|d_\Phi\|_2^2.
\label{eq:app_center_step_bound}
\end{equation}
Using $1-\kappa\alpha/2\ge1/2$ and $H_{t_{\mathrm{seg}},h}(0)=\min_{t\in[0,1]}H_{t,h}(0)$, we obtain
\begin{equation}
\min_{t\in[0,1]}H_{t,h}(0)-H_{t_{\mathrm{seg}},h}(q_\alpha)
\ge\frac{\alpha}{2}\|d_\Phi\|_2^2>0.
\label{eq:app_center_descent_gain}
\end{equation}
By Eq.~\eqref{eq:app_segment_stationarity}, $q_\alpha^\top v=0$.
Since $q_\alpha\ne0$, the displaced center lies outside the endpoint line.
Finally, $t_{\mathrm{seg}}\in(0,1)$ allows this displacement to be realized by
\begin{equation}
P_1=\frac{w_0+w_T}{2}
 +\frac{q_\alpha}{2t_{\mathrm{seg}}(1-t_{\mathrm{seg}})}.
\label{eq:app_control_point_realization}
\end{equation}
This completes the proof.
\end{proof}

The result guarantees neither global optimality nor joint Adam convergence.

For single-state matching ($h=0$), fix $\widehat X,y$, $u\ne0$, $\lambda_p>0$, and $\omega_\ell>0$.
Let $W=\operatorname{diag}(\omega_1I_{n_1},\ldots,\omega_MI_{n_M})$ and $R(q;u)=4\lambda_p q^\top Wq$, where $n_\ell$ is the size of weight or bias tensor $\ell$.
Write $D(w)=\mathcal L_{\mathrm{cos},0}(\widehat X,w)$ for Eq.~\eqref{eq:method-shared-cosine} at $h=0$, assuming nonzero model gradients.

\begin{proposition}[Local sensitivity of cosine matching]
\label{prop:cosine_sensitivity}
\upshape
For fixed $\widehat X,y,u\ne0$ and $w$, let $a=g_{\widehat X}(w+q)$ and $b=g_{\widehat X}(w)$.
Suppose $\min\{\|a\|_2,\|b\|_2\}\ge m>0$ and $\|a-b\|_2\le\sum_\ell L_\ell\|q_\ell\|_2$ for finite $L_\ell\ge0$.
Then
\begin{equation}
|D(w+q)-D(w)|\le\frac{2}{m}
 \sqrt{\sum_\ell L_\ell^2/\omega_\ell}
 \sqrt{\sum_\ell\omega_\ell\|q_\ell\|_2^2}.
\label{eq:method-sensitivity}
\end{equation}
\end{proposition}

\begin{proof}[\normalfont\bfseries Proof]
By Cauchy--Schwarz, the triangle and reverse triangle inequalities, and the norm lower bound,
\begin{equation}
|D(w+q)-D(w)|\le\frac{2}{m}\|a-b\|_2.
\label{eq:method-normalization-bound}
\end{equation}
Applying the assumed gradient bound and weighted Cauchy--Schwarz proves Eq.~\eqref{eq:method-sensitivity}.
\end{proof}
For fixed $W,L_\ell$, bounding $R$ controls the displacement factor $\sqrt{R(q;u)/(4\lambda_p)}$ in Eq.~\eqref{eq:method-sensitivity}.
The bound establishes neither optimal weights nor an advantage over uniform weights.

An optional weighting scheme sets $e_\ell=\|u_\ell\|_2^2/n_\ell$, $\bar e=\|u\|_2^2/d>0$, and $\rho_\ell=e_\ell/\bar e$, where $d=\sum_\ell n_\ell$.
These ratios are unchanged by nonzero rescaling of $u$.
For $\varepsilon>0$ and $0<a\le b$, define
\begin{equation}
\tilde\omega_\ell=\operatorname{clip}\!\left[
 \left(\frac{1+\varepsilon}{\rho_\ell+\varepsilon}\right)^{1/2},a,b\right].
\label{eq:method-raw-weight}
\end{equation}
Normalizing these coefficients gives
\begin{equation}
\omega_\ell=\frac{\tilde\omega_\ell}
 {d^{-1}\sum_jn_j\tilde\omega_j}.
\label{eq:method-normalized-weight}
\end{equation}
They are positive, satisfy $\sum_\ell n_\ell\omega_\ell=d$, and remain fixed during reconstruction.
Net-update cancellation can produce small $\rho_\ell$ despite substantial intermediate updates, so this movement prior neither estimates $L_\ell$ nor recovers the hidden trajectory.
The displacement penalty in Eq.~\eqref{eq:method-objective} corresponds to $W=I$.

\setcounter{dbltopnumber}{4}
\begin{table*}[pos=!t] \centering
\caption{Reconstruction performance on FEMNIST with MLP and ViT under two multi-step FedAvg settings. Best results are shown in bold. The NL-SME columns are highlighted in gray.} \label{tab:mlp_vit_e20n50_b10_b20}
\scriptsize \setlength{\tabcolsep}{2.6pt} \renewcommand{\arraystretch}{1.05}
\resizebox{\textwidth}{!}{%
\begin{tabular}{ll|ccccc|ccccc}
\toprule
\multirow{2}{*}{Model} & \multirow{2}{*}{Metric} & \multicolumn{5}{c|}{E20N50B10} & \multicolumn{5}{c}{E20N50B20} \\
\cmidrule(lr){3-7} \cmidrule(lr){8-12}
& & IG & SME & FEDLEAK & GI-NAS & \cellcolor{gray!20}NL-SME & IG & SME & FEDLEAK & GI-NAS & \cellcolor{gray!20}NL-SME \\
\midrule
\multirow{4}{*}{MLP} & PSNR $\uparrow$ & 23.0392 $\pm$ 0.3757 & 26.5598 $\pm$ 0.4393 & 21.3122 $\pm$ 0.1655 & 22.6927 $\pm$ 0.9357 & \cellcolor{gray!20}\textbf{32.7973 $\pm$ 1.3201} & 23.9940 $\pm$ 0.4845 & 28.5443 $\pm$ 0.7181 & 21.5046 $\pm$ 0.1279 & 24.0139 $\pm$ 1.2551 & \cellcolor{gray!20}\textbf{32.0024 $\pm$ 1.0010} \\
& SSIM $\uparrow$ & 0.7097 $\pm$ 0.0133 & 0.8472 $\pm$ 0.0199 & 0.6395 $\pm$ 0.0084 & 0.6678 $\pm$ 0.0179 & \cellcolor{gray!20}\textbf{0.8790 $\pm$ 0.0249} & 0.7459 $\pm$ 0.0108 & 0.8700 $\pm$ 0.0165 & 0.6465 $\pm$ 0.0110 & 0.7188 $\pm$ 0.0089 & \cellcolor{gray!20}\textbf{0.8723 $\pm$ 0.0225} \\
& FSIM $\uparrow$ & 0.7709 $\pm$ 0.0158 & 0.9106 $\pm$ 0.0088 & 0.7655 $\pm$ 0.0078 & 0.7282 $\pm$ 0.0192 & \cellcolor{gray!20}\textbf{0.9446 $\pm$ 0.0146} & 0.8149 $\pm$ 0.0044 & 0.9412 $\pm$ 0.0069 & 0.7786 $\pm$ 0.0098 & 0.7760 $\pm$ 0.0141 & \cellcolor{gray!20}\textbf{0.9430 $\pm$ 0.0148} \\
& LPIPS $\downarrow$ & 0.2910 $\pm$ 0.0187 & 0.1257 $\pm$ 0.0063 & 0.3391 $\pm$ 0.0196 & 0.3619 $\pm$ 0.0126 & \cellcolor{gray!20}\textbf{0.0818 $\pm$ 0.0179} & 0.2513 $\pm$ 0.0076 & \textbf{0.0798 $\pm$ 0.0080} & 0.3267 $\pm$ 0.0058 & 0.3026 $\pm$ 0.0124 & \cellcolor{gray!20}0.0853 $\pm$ 0.0178 \\
\midrule
\multirow{4}{*}{ViT} & PSNR $\uparrow$ & 13.3595 $\pm$ 0.1001 & 14.6736 $\pm$ 0.2165 & 11.2598 $\pm$ 0.3206 & 14.1503 $\pm$ 0.8305 & \cellcolor{gray!20}\textbf{22.7876 $\pm$ 1.3390} & 14.6726 $\pm$ 0.0877 & 16.5987 $\pm$ 0.2423 & 11.6627 $\pm$ 0.5387 & 15.7447 $\pm$ 0.9989 & \cellcolor{gray!20}\textbf{24.8572 $\pm$ 0.6033} \\
& SSIM $\uparrow$ & 0.3804 $\pm$ 0.0093 & 0.4422 $\pm$ 0.0143 & 0.2920 $\pm$ 0.0112 & 0.3908 $\pm$ 0.0286 & \cellcolor{gray!20}\textbf{0.8195 $\pm$ 0.0404} & 0.4335 $\pm$ 0.0115 & 0.5086 $\pm$ 0.0139 & 0.3105 $\pm$ 0.0169 & 0.4541 $\pm$ 0.0366 & \cellcolor{gray!20}\textbf{0.8767 $\pm$ 0.0165} \\
& FSIM $\uparrow$ & 0.4290 $\pm$ 0.0063 & 0.4507 $\pm$ 0.0071 & 0.3994 $\pm$ 0.0082 & 0.4641 $\pm$ 0.0139 & \cellcolor{gray!20}\textbf{0.7648 $\pm$ 0.0448} & 0.4531 $\pm$ 0.0053 & 0.4941 $\pm$ 0.0090 & 0.4083 $\pm$ 0.0086 & 0.4981 $\pm$ 0.0189 & \cellcolor{gray!20}\textbf{0.8290 $\pm$ 0.0168} \\
& LPIPS $\downarrow$ & 0.6152 $\pm$ 0.0049 & 0.5839 $\pm$ 0.0074 & 0.6515 $\pm$ 0.0060 & 0.5778 $\pm$ 0.0352 & \cellcolor{gray!20}\textbf{0.2491 $\pm$ 0.0359} & 0.5824 $\pm$ 0.0042 & 0.5295 $\pm$ 0.0098 & 0.6348 $\pm$ 0.0103 & 0.5406 $\pm$ 0.0365 & \cellcolor{gray!20}\textbf{0.1885 $\pm$ 0.0144} \\
\bottomrule
\end{tabular}%
}
\end{table*}

\begin{table*}[width=0.485\textwidth,pos=!t] \centering
\begin{minipage}[t]{0.485\textwidth}
\vspace{0pt}
\centering
\caption{Reconstruction performance on Tiny-ImageNet with a three-layer MLP under $T=100$.} \label{tab:tiny_imagenet_t100_summary}
\scriptsize \setlength{\tabcolsep}{2.2pt} \renewcommand{\arraystretch}{1.05}
\resizebox{\linewidth}{!}{%
\begin{tabular}{lcccccc}
\toprule
\textbf{Metric} & \textbf{IG} & \textbf{SME} & \textbf{FEDLEAK} & \textbf{GI-NAS} & \textbf{DGGI} & \cellcolor{gray!20}\textbf{NL-SME} \\
\midrule
PSNR $\uparrow$ & 14.4599 $\pm$ 0.9715 & 16.1264 $\pm$ 2.0827 & 12.0957 $\pm$ 0.9456 & 11.9358 $\pm$ 1.0546 & 16.0336 $\pm$ 0.2064 & \cellcolor{gray!20}\textbf{26.6640 $\pm$ 0.9827} \\
SSIM $\uparrow$ & 0.4966 $\pm$ 0.0604 & 0.5050 $\pm$ 0.1115 & 0.3342 $\pm$ 0.0485 & 0.3035 $\pm$ 0.0769 & 0.5026 $\pm$ 0.0442 & \cellcolor{gray!20}\textbf{0.8634 $\pm$ 0.0235} \\
FSIM $\uparrow$ & 0.7469 $\pm$ 0.0319 & 0.7683 $\pm$ 0.0577 & 0.6766 $\pm$ 0.0237 & 0.5998 $\pm$ 0.0367 & 0.7647 $\pm$ 0.0207 & \cellcolor{gray!20}\textbf{0.9216 $\pm$ 0.0123} \\
LPIPS $\downarrow$ & 0.3872 $\pm$ 0.0473 & 0.4248 $\pm$ 0.1094 & 0.5357 $\pm$ 0.0423 & 0.5956 $\pm$ 0.0578 & 0.4594 $\pm$ 0.0526 & \cellcolor{gray!20}\textbf{0.1403 $\pm$ 0.0246} \\
\bottomrule
\end{tabular}%
}
\end{minipage}
\hfill
\begin{minipage}[t]{0.485\textwidth}
\vspace{0pt}
\centering
\caption{Reconstruction performance on OrganAMNIST64 with CNNmnist under $E=20$, $N=20$, and $B=10$.} \label{tab:organamnist64_e20n20b10}
\scriptsize \setlength{\tabcolsep}{2.4pt} \renewcommand{\arraystretch}{1.05}
\resizebox{\linewidth}{!}{%
\begin{tabular}{lcccccc}
\toprule
\textbf{Metric} & \textbf{IG} & \textbf{SME} & \textbf{FEDLEAK} & \textbf{GI-NAS} & \textbf{DGGI} & \cellcolor{gray!20}\textbf{NL-SME} \\
\midrule
PSNR $\uparrow$ & 12.7446 $\pm$ 1.4258 & 17.2748 $\pm$ 0.9291 & 12.5169 $\pm$ 1.3316 & 13.1249 $\pm$ 1.6183 & 17.6392 $\pm$ 0.7409 & \cellcolor{gray!20}\textbf{18.5320 $\pm$ 1.4375} \\
SSIM $\uparrow$ & 0.2837 $\pm$ 0.0515 & 0.5252 $\pm$ 0.0313 & 0.2685 $\pm$ 0.0550 & 0.3159 $\pm$ 0.0593 & 0.5463 $\pm$ 0.0278 & \cellcolor{gray!20}\textbf{0.5719 $\pm$ 0.0476} \\
FSIM $\uparrow$ & 0.6339 $\pm$ 0.0355 & 0.7247 $\pm$ 0.0267 & 0.6271 $\pm$ 0.0339 & 0.6507 $\pm$ 0.0420 & 0.7333 $\pm$ 0.0253 & \cellcolor{gray!20}\textbf{0.7462 $\pm$ 0.0368} \\
LPIPS $\downarrow$ & 0.5717 $\pm$ 0.0122 & 0.4680 $\pm$ 0.0159 & 0.5768 $\pm$ 0.0139 & 0.5437 $\pm$ 0.0260 & 0.4603 $\pm$ 0.0167 & \cellcolor{gray!20}\textbf{0.4486 $\pm$ 0.0385} \\
\bottomrule
\end{tabular}%
}
\end{minipage}
\end{table*}

\begin{figure*}[pos=!t] \centering
\begin{minipage}[t]{0.485\textwidth} \centering \includegraphics[width=\linewidth]{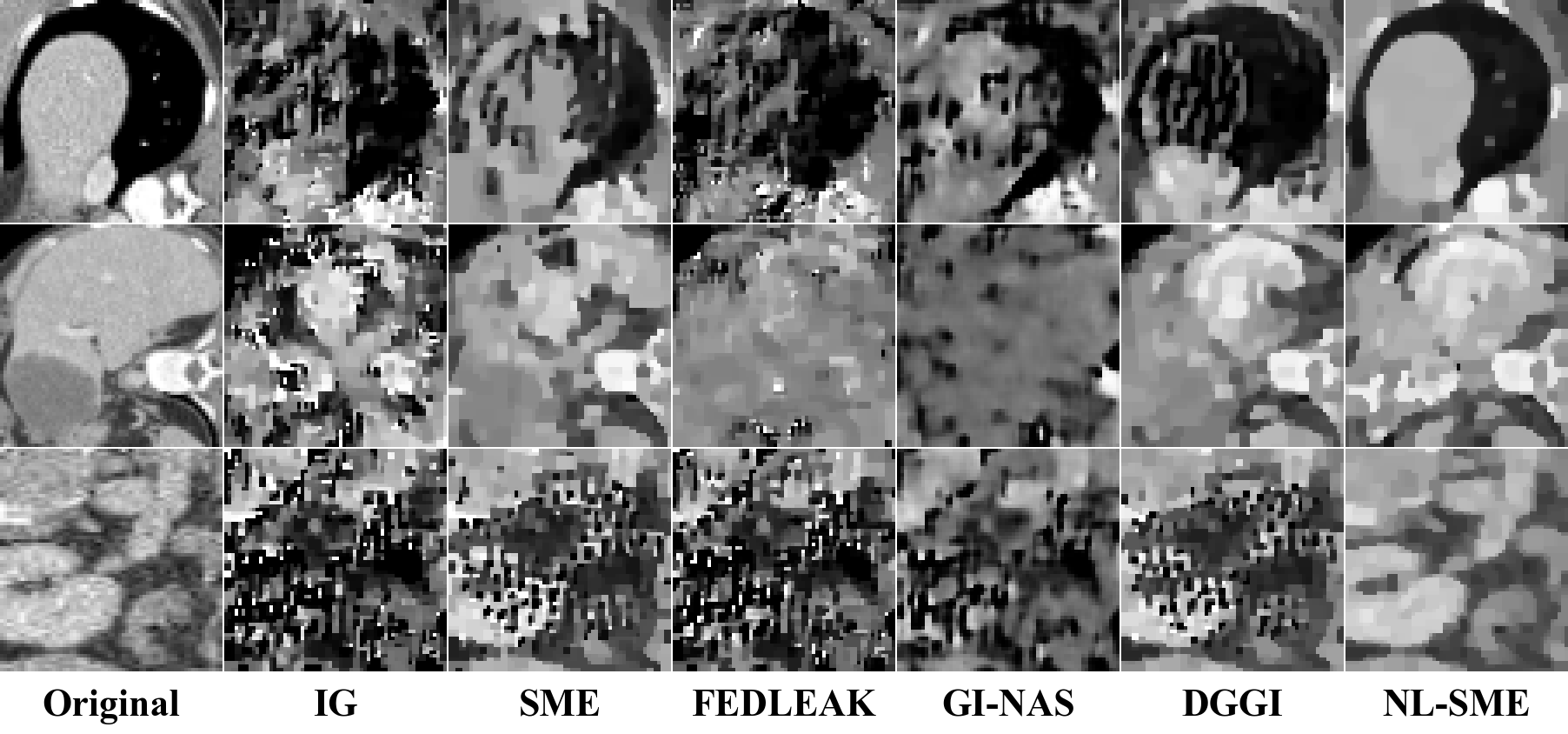} \vspace{1mm} {\footnotesize \textbf{(a)} OrganAMNIST64\par} \end{minipage}
\hfill
\begin{minipage}[t]{0.485\textwidth} \centering \includegraphics[width=\linewidth]{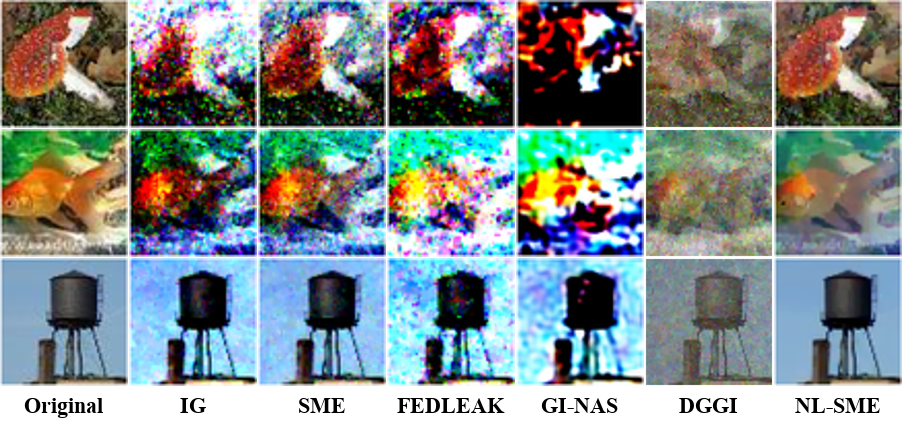} \vspace{1mm} {\footnotesize \textbf{(b)} Tiny-ImageNet\par} \end{minipage}
\caption{Reconstructed images from different gradient inversion attacks on OrganAMNIST64 and Tiny-ImageNet.} \label{fig:qualitative_organ_tiny}
\end{figure*}

\begin{figure*}[pos=!t] \centering
\begin{subfigure}[t]{0.24\textwidth} \centering \includegraphics[width=\linewidth]{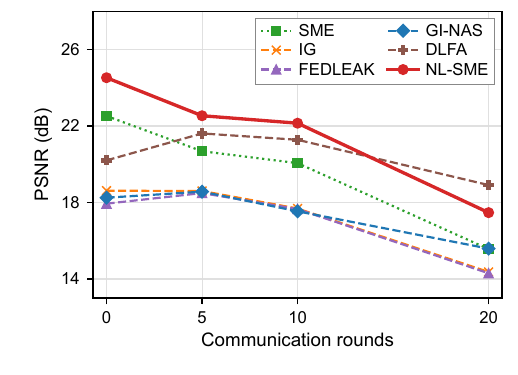} \caption{PSNR $\uparrow$} \label{fig:trained_network_psnr} \end{subfigure}
\hfill
\begin{subfigure}[t]{0.24\textwidth} \centering \includegraphics[width=\linewidth]{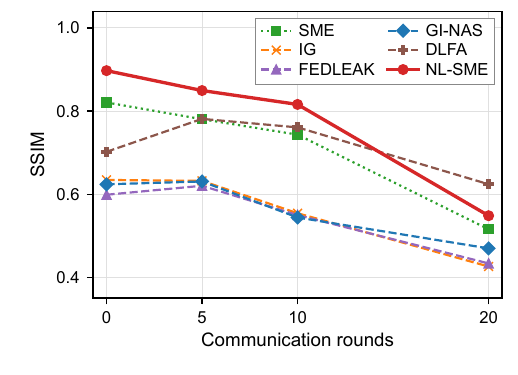} \caption{SSIM $\uparrow$} \label{fig:trained_network_ssim} \end{subfigure}
\hfill
\begin{subfigure}[t]{0.24\textwidth} \centering \includegraphics[width=\linewidth]{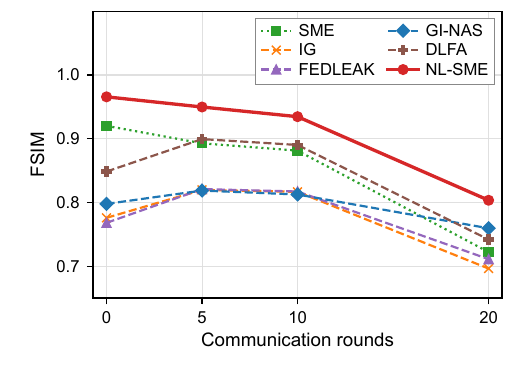} \caption{FSIM $\uparrow$} \label{fig:trained_network_fsim} \end{subfigure}
\hfill
\begin{subfigure}[t]{0.24\textwidth} \centering \includegraphics[width=\linewidth]{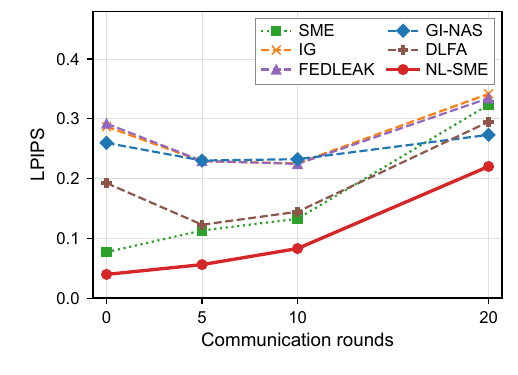} \caption{LPIPS $\downarrow$} \label{fig:trained_network_lpips} \end{subfigure}
\caption{Reconstruction performance across training stages on FEMNIST with $E=10$, $N=50$, and $B=10$. Each point averages 50 images from one run, with $R=0$ denoting random initialization.} \label{fig:trained_network_metrics}
\end{figure*}

\section{Additional results}
\label{appendix:additional_results}
\textbf{Evaluation on MLP and ViT.}
Table~\ref{tab:mlp_vit_e20n50_b10_b20} reports the reconstruction performance on FEMNIST with MLP and ViT at batch sizes of 10 and 20.
NL-SME achieves the highest mean PSNR, SSIM, and FSIM for both architectures at both batch sizes.
For LPIPS, NL-SME achieves the lowest mean value in three settings, while SME performs best on MLP with a batch size of 20.
These results support the effectiveness of NL-SME for reconstruction from multi-step updates under these model settings.
The ViT uses a single image patch, and evaluation on larger or multi-patch transformers is left for future work.

\textbf{Attack results on medical images.}
Table~\ref{tab:organamnist64_e20n20b10} reports the reconstruction performance of six attack methods from multi-step updates on OrganAMNIST64.
NL-SME achieves the best mean performance across all four metrics in this setting.
For instance, NL-SME achieves a mean PSNR of \(18.53\) dB, compared with \(17.64\) dB for DGGI.

\textbf{Evaluation on natural images.}
Table~\ref{tab:tiny_imagenet_t100_summary} reports reconstruction results on Tiny-ImageNet in a three-layer MLP sanity-check setting.
NL-SME achieves the best mean performance across all four metrics in this setting.
For instance, NL-SME achieves a mean PSNR of \(26.66\) dB, compared with \(16.13\) dB for SME.

\textbf{Evaluation on trained networks.}
Following SME~\cite{zhu2023surrogate}, we evaluate reconstruction on FEMNIST with CNNmnist at different training stages.
The model is trained by FedAvg with the attacked client excluded.
All methods use the same private data, fixed ground-truth labels, and model endpoints at each stage, with $1000$ reconstruction iterations but different computational costs.
For DLFA, we use a PyTorch adaptation of the full-input attack
with an order-invariant prior and adopt Adam with an initial
learning rate of $0.4$.
For all methods in this evaluation, we obtain a one-to-one assignment
by minimizing total pixel-wise MSE without label constraints and
use the same assignment for all four metrics.
As shown in Figure~\ref{fig:trained_network_metrics}, NL-SME achieves the best values across all four metrics at initialization and after $5$ and $10$ rounds.
After $20$ rounds, DLFA~\cite{dimitrov2022data} achieves higher PSNR and SSIM than NL-SME, while NL-SME retains the highest FSIM and lowest LPIPS.
For this client and training trajectory, all six methods yield poorer reconstruction after $20$ rounds than at initialization.

\FloatBarrier

\bibliographystyle{elsarticle-num}
\bibliography{references}

\end{document}